%% file: arxiv2025.tex
\documentclass{article} 
\usepackage[dvipsnames]{xcolor}
\usepackage[preprint]{colm2025_conference}
\usepackage{microtype}
\usepackage{graphicx}
\usepackage{wrapfig}
\usepackage{caption}
\usepackage{subcaption}
\usepackage{microtype}
\usepackage{hyperref}
\usepackage{url}
\usepackage{booktabs}
\usepackage{amsmath}
\usepackage{listings}
\usepackage{lipsum}
\usepackage{algorithm}
\usepackage{algpseudocode}
\usepackage{color, colortbl}
\definecolor{codegreen}{rgb}{0,0.6,0}
\definecolor{codegray}{rgb}{0.5,0.5,0.5}
\definecolor{codepurple}{rgb}{0.58,0,0.82}
\definecolor{backcolour}{rgb}{0.95,0.95,0.92}
\definecolor{DarkPink}{rgb}{0.5,0.0,0.18}
\definecolor{DarkGreen}{rgb}{0.1,0.5,0.1}
\definecolor{DarkRed}{rgb}{0.76,0.3,0.15}
\definecolor{DarkBlue}{rgb}{0.1,0.1,0.7}
\definecolor{DarkYellow}{rgb}{.79,.79,0}
\usepackage{lineno}
\usepackage{adjustbox}
\definecolor{darkblue}{rgb}{0, 0, 0.5}
\hypersetup{colorlinks=true, citecolor=darkblue, linkcolor=darkblue, urlcolor=darkblue}
\input{math_commands.tex}

\usepackage{enumitem}
\newcommand{\lt}{\ensuremath <}

\title{NeuralGrok: \\Accelerate Grokking by Neural Gradient Transformation}

\newcommand{\equalcontrib}{\textsuperscript{\dag}}

\author{{Xinyu Zhou\equalcontrib}, Simin Fan\equalcontrib, Martin Jaggi \\
EPFL\\
\texttt{firstname.lastname@epfl.ch} \\
\And
Jie Fu\textsuperscript{*} \\
Shanghai AI Lab \\
\texttt{fujie@pjlab.org.cn}
}
\begin{document}

\ifcolmsubmission
\linenumbers
\fi

\maketitle
{\renewcommand{\thefootnote}{}
\footnotetext{\textsuperscript{\dag}These authors contributed equally to this work. Xinyu did this work during the internship at Shanghai AI Lab.}\footnotetext{\textsuperscript{*}The corresponding author. }}

\begin{abstract}
\textit{Grokking} is proposed and widely studied as an intricate phenomenon in which generalization is achieved after a long-lasting period of overfitting.
In this work, we propose \textsc{NeuralGrok}, a novel gradient-based approach that learns an optimal gradient transformation to accelerate the generalization of transformers in arithmetic tasks. 
Specifically, \textsc{NeuralGrok} trains an auxiliary module (e.g., an MLP block) in conjunction with the base model. This module dynamically modulates the influence of individual gradient components based on their contribution to generalization, guided by a bilevel optimization algorithm.
Our extensive experiments demonstrate that \textsc{NeuralGrok} significantly accelerates generalization, particularly in challenging arithmetic tasks. We also show that \textsc{NeuralGrok} promotes a more stable training paradigm, constantly reducing the model's complexity, while traditional regularization methods, such as weight decay, can introduce substantial instability and impede generalization. We further investigate the intrinsic model complexity leveraging a novel \textbf{A}bsolute \textbf{G}radient \textbf{E}ntropy (AGE) metric, which explains that \textsc{NeuralGrok} effectively facilitates generalization by reducing the model complexity.
We offer valuable insights on the grokking phenomenon of Transformer models, which encourages a deeper understanding of the fundamental principles governing generalization ability. We provide the code in \url{https://github.com/Blackzxy/NeuralOptGrok/tree/neuralgrad}.
\end{abstract}

\section{Introduction}
Understanding the generalization mechanism of over-parameterized neural networks is a long-standing challenge in the field of deep learning. 
\citet{power2022grokking} observed an intriguing phenomenon termed \textit{Grokking}, wherein a transformer model exhibits delayed generalization on unseen data long after overfitting to the training data on a simple arithmetic task.  
Numerous investigations have sought to understand and justify this phenomenon from a representation learning perspective \citep{liu2022understandinggrokkingeffectivetheory,kumar2024grokkingtransitionlazyrich,fan2024deepgrokkingdeepneural} and theoretical analysis\citep{davies2023unifyinggrokkingdoubledescent,thilak2022slingshotmechanismempiricalstudy,prieto2025grokkingedgenumericalstability,humayun2024deepnetworksgrok}. Recently, \citet{lee2024grokfastacceleratedgrokkingamplifying} demonstrated that by amplifying the low-frequency component of the gradient by a low-pass filter (LPF), the generalization can be greatly accelerated.

Instead of strict low-pass filtering, we propose \textsc{NeuralGrok}, a bilevel algorithm which trains an adaptive and learnable gradient transformation pattern to accelerate generalization under the grokking phenomenon. 
Specifically, we train an auxiliary module termed \textit{neural-amplifier}, implemented as a simple MLP block, in conjunction with the base model. 
This module dynamically modulates the influence of individual gradient components based on their contribution to generalization, guided by a bilevel optimization algorithm. 
In the inner loop, the model gradients are first tuned by the \textit{neural-amplifier} and then applied to update the model parameters; In the outer loop, the \textit{neural-amplifier} is trained to minimize stochastic loss from a separated validation set. In our implementation, the validation set is a small subset of the original training set. Conceptually, the \textit{neural-amplifier} is trained to minimize the generalization gap \citep{johnson2023inconsistencyinstabilitygeneralizationgap}, effectively transforming the gradient to facilitate the learning of generalizable features.

Through extensive experiments on arithmetical tasks, we demonstrate that \textsc{NeuralGrok} significantly accelerates generalization, ranging from simple operations (e.g., '+, -, $\times$') to complex and composite arithmetic tasks. 
In addition, compared to commonly used regularization such as \textit{weight-decay}, we further show that the gradient transformation paradigm adopted by \textsc{NeuralGrok} yields a more stable generalization behavior, while applying \textit{weight-decay} can introduce substantial instability and impede generalization. We further investigate into the intrinsic complexity of the model leveraging the \textbf{absolute weight entropy}~\citep{golechha2024progressmeasuresgrokkingrealworld} over training steps, which explains \textsc{NeuralGrok} effectively stabilizes the training and shortens the phase transition from memorization and generalization.

We aim to address the following research questions in subsequent sections:
\begin{itemize} 
\setlength{\itemindent}{-2.7em}
    \item \textbf{RQ1:} Could a simple auxiliary neural network effectively learn a gradient transformation that accelerates the generalization of the base model?
    \vspace{0.2em}
    \item \textbf{RQ2:} Does the gradient transformation method lead to a stable generalization pattern? How does it compare to traditional regularization approaches, such as weight-decay?
    \vspace{0.2em}
    \item \textbf{RQ3:} What is the intrinsic mechanism that can interpret the phase transition from memorization to generalization?
\end{itemize}

\section{\textsc{NeuralGrok}: Accelerate Generalization by Learnable Gradient Transformation}\label{sec:neuralgrok}
\paragraph{Learning Generalizable Gradients by Bilevel Optimization.}
We hereby introduce the pipeline of \textsc{NeuralGrok}. 
Alongside the standard training run, we train an auxiliary \textit{neural-amplifier} $G(\varphi)$ to learn a gradient transformation pattern that enhances the generalization capabilities of the base model $M(\vtheta)$.

We formulate the learning of gradient patterns as a bilevel optimization problem: 
\begin{equation}\label{equ:bilevel}
    \vtheta \in \argmin_{\vtheta} L(\vtheta, \varphi^\star, \mathcal{D}_{inner}) \qquad \text{s.t.} \quad \varphi^{\star} \in \argmin_{\varphi} L(\vtheta, \varphi, \mathcal{D}_{outer}) 
\end{equation}
Given a partition of the training data $\mathcal{D}^{train} = \{\mathcal{D}_{inner}, \mathcal{D}_{outer}\}$, we optimize the base transformer model on $\mathcal{D}_{inner}$ while simultaneously tuning the \textit{neural-amplifier} on $\mathcal{D}_{outer}$. In the inner loop, we compute the original model gradients $\vg$ on $\mathcal{D}_{inner}$ then apply the \textit{neural-amplifier} to transform these gradients. The transformed gradients $\vg'$ are used to update the base model $M(\vtheta)$.  
In the outer loop, we freeze the base model parameters while optimizing the \textit{neural-amplifier} to minimize the same next-word prediction cross-entropy loss on $\mathcal{D}_{outer}$. Since the updats on the transformer model directly depends on the gradient transformation, the loss $\mathcal{D}_{outer}$ from the updated model is also associated to the \textit{neural-amplifier}, parametrized with $\varphi$. We present the \texttt{Learn-Amplifier} function in Algorithm \ref{alg:neuralgrok-outer}. We update the base model in each inner-loop for $T$ steps before conducting the outer-loop step.
Throughout training, we monitor both the accuracy on the training set and a held-out test set. Ideally, the \textit{neural-amplifier} could foster the base model to learn more generalizable features, thereby reducing the gap between overfitting, where the model merely memorizes the training data, and generalization, where the model effectively extrapolates to unseen examples in the test set. We provide the complete bilevel \textsc{NeuralGrok} algorithm in Algorithm \ref{alg:neuralgrok}.

\paragraph{Model Architectures.} In our experiments, we apply a decoder-only transformer as $M(\vtheta)$, with a simple MLP block as the \textit{neural-amplifier} $G$, parametrized by $\varphi$, mapping the main parameters $\vtheta$ (or their gradients) to the same space. 
The \textit{neural-amplifier} $G(\varphi)$ is described as a probability distribution $\vp$ over all gradient entries to show modulate influence. Subsequently, we apply a rescaling to constrain the gradient magnitude to be a constant $c$. Specifically, given an original model gradient $\vg$, $G(\varphi)$ applies the following transformation to get a modulated gradient $\vg'$.
\begin{align}
    \vp =\texttt{softmax}\left (\texttt{MLP}_{\varphi}(\vg) \right ), \qquad \vg' = c \cdot \frac{\vp\cdot \vg}{\| \vp\cdot \vg\|_2} 
    \label{equ:neuralgrok}
\end{align}
With a probability distribution $\vp\in \Delta^{|\vg|}$, the neural-amplifier applies a rotation on the original gradient $\vg$ without changing its magnitude, while the rescaling coefficient $c$ modifies the scale of the gradient. Note that $c$ is not learnable in the current framework. 
If without specification, we apply a constant $c=1.0$ as the standard gradient normalization in our experiments. 
We provide more implementation details on the \textit{neural-amplifier} in Appendix \ref{appd:mlp}.

\begin{algorithm}[!h]
\caption{\textsc{NeuralGrok}}\label{alg:neuralgrok}
\begin{algorithmic}
\Require \small Given a partition of the training set $\mathcal{D}^{train}$=$\{\mathcal{D}_{inner}, \mathcal{D}_{outer} \}$, base model model $M(\vtheta)$ with optimizer $\texttt{Opt}_M$, \textit{neural-amplifier} $G(\varphi)$ with meta optimizer $\texttt{Opt}_G$, and inner-loop frequency $T$. The learning rate at step $t$ is given by $\eta_{\vtheta,t}$, $\eta_{\varphi,t}$. We also have the access to the stochastic loss function $L(\vtheta, \mathcal{D})$ and \texttt{Learn-Amplifier}$(\varphi, \texttt{Opt}_G, \mathcal{D}_{outer})$ function to optimize $G(\varphi)$.
\\
\vspace{1.5mm}
\textbf{Init:} $t\leftarrow 0$, $\vtheta\leftarrow \vtheta_0$, $\varphi\leftarrow \varphi_0$
\vspace{1.5mm}
\While{$\vtheta_t$ is not converged}
\vspace{1.5mm}
\State \textcolor{codegray}{\footnotesize \#  Inner-loop: train base model $M$}
\State{Sample $B_t\subset \mathcal{D}_{inner}$}
\State $\vg_t = \nabla_{\vtheta} \mathcal{L}(\vtheta_t, B_t)$ \hspace{2.7em} \textcolor{codegray}{\footnotesize \# Get model's stochastic gradients $\vg$}
\State  $\displaystyle \vg'_t=G(\vg_t, \varphi_t)$ \hspace{4em} \textcolor{codegray}{\footnotesize \# Transform gradients}
\State $\displaystyle \vtheta_{t+1} \leftarrow \texttt{Opt}_M(\vtheta_t, \vg_t', \eta_{\vtheta,t})$ \hspace{1.2em} \textcolor{codegray}{\footnotesize \# Optimize model with new gradients $\vg_t'$}
\\
\If{$t \% T = 0$}
\vspace{1.5mm}
\State \textcolor{codegray}{\footnotesize \# Outer-loop: optimize \textit{neural-amplifier}}
\State $\displaystyle \varphi_{t+1} \leftarrow$ \texttt{Learn-Amplifier}$(\varphi_t, \texttt{Opt}_G, \mathcal{D}_{outer}, \eta_{\varphi,t})$ 
\vspace{1.5mm}
\EndIf
\State{$t\leftarrow t+1$}
\EndWhile

\end{algorithmic}
\end{algorithm}

\begin{algorithm}[!h]
\caption{\texttt{Learn-Amplifier}}\label{alg:neuralgrok-outer}
\begin{algorithmic}
\Require \small Given the training set $\mathcal{D}^{train}$=$\{\mathcal{D}_{inner}, \mathcal{D}_{outer} \}$, the \textit{neural-amplifier} $G(\varphi)$, the meta-optimizer $\texttt{Opt}_G$ and a copy of base model $M'(\vtheta)$. We also have the learning rate $\eta_{\vtheta}$, $\eta_{\varphi}$, and the loss function $L(\vtheta, D)$.
\\
\State \textcolor{codegray}{\footnotesize \#  Update copied base model $M'$ with $G(\varphi)$ on mini-batch $\mathcal{B}_{inner}$}
\State $\vg_{\vtheta} = \nabla_{\vtheta} \mathcal{L}(\vtheta, \mathcal{B}_{inner})$ \quad \textcolor{codegray}{\footnotesize \# Get model's gradients $\vg_{\vtheta}$}
\State  $\displaystyle \vg'_{\vtheta}=G(\varphi, \vg_{\vtheta})$ \qquad \hspace{1.5em} \textcolor{codegray}{\footnotesize \# Transform gradients}
\State $\displaystyle \vtheta' \leftarrow \vtheta - \eta_{\vtheta} \vg'_{\vtheta}$ \qquad \qquad \textcolor{codegray}{\footnotesize \# Optimize model with SGD on new gradients $\vg'_{\vtheta}$}
\\
\State \textcolor{codegray}{\footnotesize \# Optimize \textit{neural-amplifier}}
\State $\vg_{\varphi} = \nabla_{\varphi} \mathcal{L}(\vtheta', \mathcal{D}_{outer}) = \nabla_{\varphi} \mathcal{L}(\vtheta - \eta_{\vtheta} G(\varphi, \vg_{\vtheta}), \mathcal{D}_{outer})$ \quad \textcolor{codegray}{\footnotesize \# Evaluate updated $M'(\vtheta')$ on $\mathcal{D}_{outer}$}
\State $\displaystyle \varphi \leftarrow \texttt{Opt}_G(\varphi, \vg_{\varphi}, \eta_{\varphi})$ 
\\
\Return{$\varphi$}
\end{algorithmic}
\end{algorithm}

\section{Experiments}
\paragraph{Arithmetic tasks.} We test \textsc{NeuralGrok} on a set of arithmetic tasks following \citet{power2022grokking} and \citet{lee2024grokfastacceleratedgrokkingamplifying}, with various difficulty levels by composing the arithmetic operations. 
Each task dataset consists of textual sequences of a mathematical equation. The simplest task is of the form $a \circ b = r$, where $a$, $b$ are input variables, $\circ$ is a binary operand and $r$ is the result. We can create a more complex task by compositional operations on $k$ input numbers and $k-1$ operands, which are defined in the form $v_1 \circ_1 v_2 \circ_2 \hdots \circ_{k-1} v_{k} = r$. We present each sequence in the tokenized form of $\langle v_1\rangle\langle v_2\rangle...\langle v_k\rangle \langle op_1\rangle \langle op_2\rangle...\langle op_{k-1}\rangle \langle =\rangle \langle r \rangle$, where $\langle x\rangle$ stands for the token corresponding to the element $x$.

Following \citet{power2022grokking} and \citet{lee2024grokfastacceleratedgrokkingamplifying}, we randomly split the whole dataset into $50\%, 50\%$ partitions into a training $\mathcal{D}^{train}$ and test set $\mathcal{D}^{test}$. For \textsc{NeuralGrok}, we further divide $\mathcal{D}^{train}$ into $\mathcal{D}_{inner}$ and $\mathcal{D}_{outer}$ with a ratio of $49:1$. For all baseline methods, the transformer model is trained on $\mathcal{D}^{train}$, while \textsc{NeuralGrok} is trained on $\mathcal{D}_{inner}$ and  $\mathcal{D}_{outer}$ following the bilevel algorithm described in \S~\ref{sec:neuralgrok}. All methods are tested on the same test set $\mathcal{D}^{test}$, which ensures a fair comparison. 
Without specification, we apply \textit{weight-decay} $wd=1e^{-3}$ as default for all experiments, since it elicits stable and balanced generalization performance on the baseline methods across various tasks. We provide the justifications on the baseline selection as follows.

\textbf{Baselines.}
We compare \textsc{NeuralGrok} with two baseline methods: (1) \textbf{Standard training}: we apply the standard autoregressive training with weight-decay; and (2) \textbf{\textsc{GrokFast}-MA} and \textbf{\textsc{GrokFast}-EMA} \citep{lee2024grokfastacceleratedgrokkingamplifying}: the transformer model is updated with average or exponential-moving average gradients from a specific window of steps. For all methods, we keep the hyperparameters (e.g., learning rate and weight-decay) constant.
We find that standard training can hardly generalize with a large weight-decay ($wd=0.01$) and \textsc{GrokFast-MA} is quite sensitive to hyperparameter settings, which can be task-dependent. 
We therefore set a constant weight-decay $wd=1e^{-3}$ across all the methods. For other hyperparameters on \textsc{GrokFast}, we follow the optimal setting as illustrated in the original paper \citep{lee2024grokfastacceleratedgrokkingamplifying}. We provide more justifications for the baselines in \autoref{appd:baseline}.

\subsection{\textsc{NeuralGrok} Accelerates Model Generalization}
We demonstrate that \textsc{NeuralGrok} effectively accelerates grokking across all arithmetic tasks compared to standard training and \textsc{GrokFast} baselines. 
Since the dynamics of \textsc{GrokFast}-EMA is unstable during training, we only include the curves of \textsc{GrokFast}-MA for comparison. We provide the complete results on \textsc{GrokFast}-EMA in \autoref{appd:baseline}.
We present the minimal optimization steps required to achieve $95\%$ test accuracy in \autoref{table:min-opt-steps}.
\begin{wrapfigure}{r}{0.55\textwidth}
  \centering  
  \begin{subfigure}[t]{0.3\linewidth}\vskip 0pt
  \includegraphics[width=\textwidth,clip]{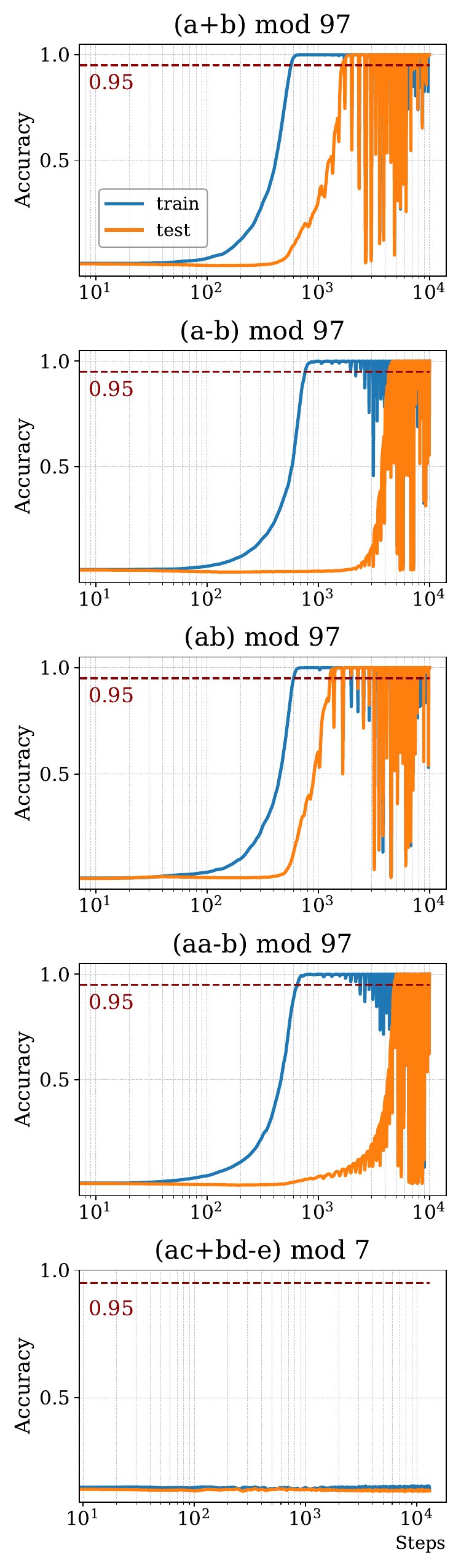}
  \caption{\small Standard}\label{fig:all_tasks_std}
  
  \end{subfigure}\hfill
  \begin{subfigure}[t]{0.3\linewidth}\vskip 0pt
  \includegraphics[width=\textwidth,clip]{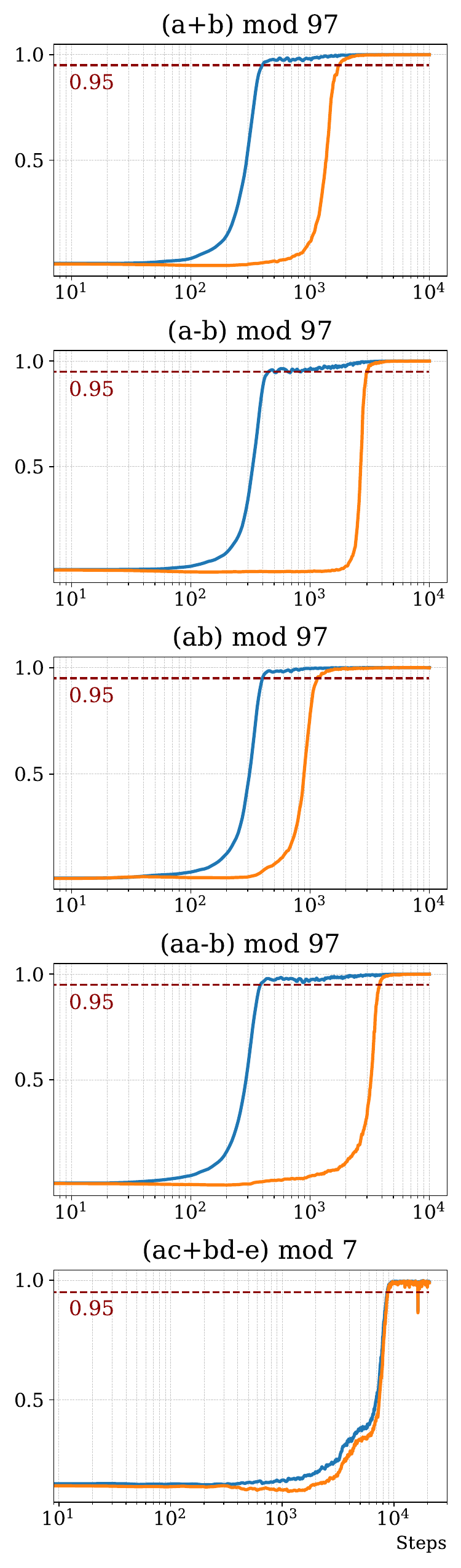}
  \caption{\small Grokfast-MA}\label{fig:all_tasks_ma}
  
  \end{subfigure}\hfill
  \begin{subfigure}[t]{0.3\linewidth}\vskip 0pt
  \includegraphics[width=\textwidth,clip]{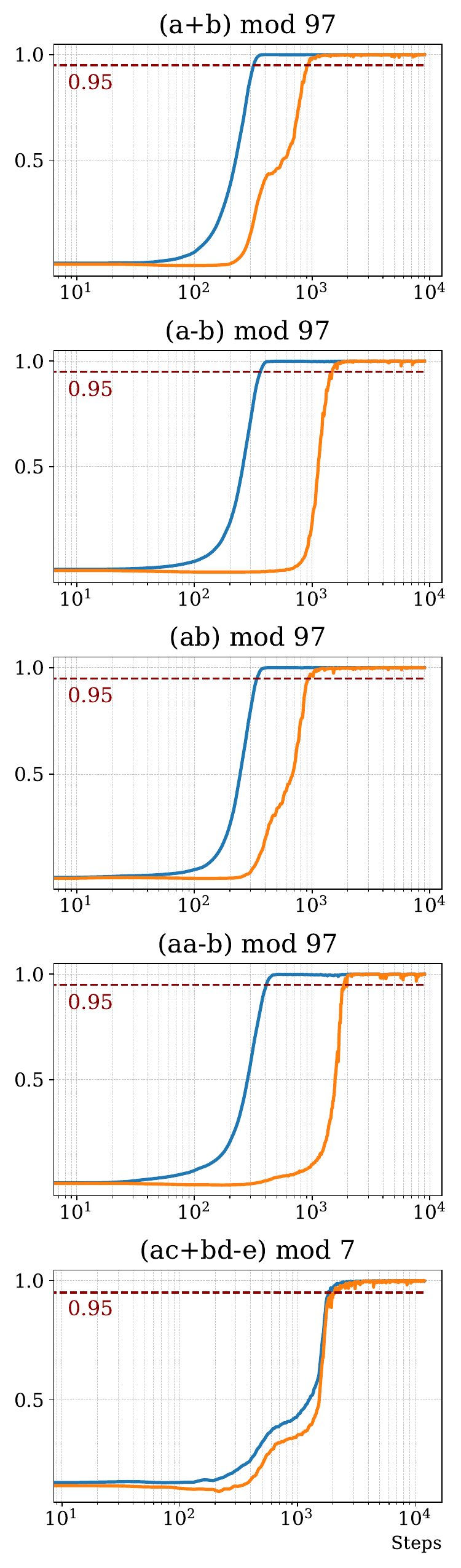}
  \caption{\small NeuralGrok}\label{fig:all_tasks_neuralgrok}
  \end{subfigure}
  \vspace{-0.5em}
  \caption{\textbf{Train and Test accuracies on arithmetic tasks.} \textsc{NeuralGrok} consistently accelerates generalization under the grokking phenomenon, especially on the challenging task.}
 \label{fig:all_tasks}
 \vspace{-2em}
\end{wrapfigure}

\textbf{Setup.} We construct five arithmetic tasks with various difficulty levels, including four tasks between two arguments: \texttt{(a+b) mod 97}, \texttt{(a-b) mod 97}, \texttt{(a$\times$b) mod 97}, \texttt{(a$\times$a-b) mod 97}, and one challenging task with five arguments: \texttt{(a$\times$c+b$\times$d-e) mod~7}. For the first four tasks, we apply a 2-layer transformer \citep{vaswani2023attentionneed} as the base model $M(\vtheta)$ with a 3-layer MLP as the neural-amplifier. We update the neural-amplifier every $T=4$ steps.
For the complicated task \texttt{(a$\times$c+b$\times$d-e) mod 7}, we adopt a 4-layer transformer as the base model. To enable a fast adaptation of the neural-amplifier, we update it every $T=1$ steps.

\textbf{Results.} We present the evolution of training and test accuracies on all five arithmetic tasks in \autoref{fig:all_tasks}. On the simple arithmetic operations with only two arguements, \textsc{NeuralGrok} obtains an acceleration in generalization up to $2.95\times$ and $2.08\times$ compared to standard training, and \textsc{GrokFast}-MA, respectively. Notably, \textsc{NeuralGrok} successfully acquires the most challenging task \texttt{(ac+bd-e) mod 7} with $4.67\times$ acceleration upon \textsc{GrokFast}-MA, while both \textsc{GrokFast}-EMA and the standard training fail to memorize nor generalize on the task within $10^6$ optimization steps. It demonstrates that the \textit{neural-amplifier} effectively learns a gradient transformation pattern, which facilitates the generalization of the base transformer model. 

\textbf{Stability of the Generalization Pattern.}
While applying standard training is able to achieve perfect test accuracy with weight-decay regularization, we find that the dynamics after generalization are extremely unstable. Across all arithmetic tasks, the test accuracy damps between the perfect score ($\sim 100\%$) and a collapsed pattern ($\lt 5\%$). 
As a conventional regularization technique, applying a larger value of weight-decay fails to help. As shown in \autoref{appd:fig:std-training-wd=0.01}, with $10\times$ larger weight-decay, the transformer model stops learning from the task, neither memorizing the training or generalizing to the test set. 
A similar damping phenomenon is also observed on \textsc{GrokFast}-EMA (\autoref{appd:fig:grokfast-ema}), which indicates the catastrophic instability in their generalization phase. 
In contrast, both \textsc{NeuralGrok} and \textsc{GrokFast}-MA exhibits superior stability in both memorization and generalization phases.

\textbf{Learnability of arithmetic tasks by transformers.}
In human-level cognition, the modular operation with basic mathematical operators $+$,$-$ are supposed to be simpler than \texttt{$\times$} and more advanced tasks with composite operators. However, most of the algorithms agree that the subtraction ($-$) operator is more challenging to learn than $+$ and \texttt{$\times$} in term of the generalization efficiency (\autoref{table:min-opt-steps}). It reflects that the human evaluated or heuristic-based difficulty levels may not applied on neural network learners, which motivates a model-based mechanistic interpretation of generalization, particularly, under the grokking phenomenon. 

\begin{table}[ht!]
\caption{\small \textbf{Minimal optimization steps needed for the model to achieve $95\%$ test accuracy.} The best results are marked in \textbf{Bold}. \textsc{NeuralGrok} consistently outperforms Standard training and \textsc{GrokFast}-MA across all the tasks.}
\vspace{-0.5em}
\label{table:min-opt-steps}
\centering
\small 
\begin{adjustbox}{max width=0.99\textwidth}
 \begin{tabular}{l| c c  c c || c} 
 \toprule
 Arithmetic Tasks & Standard & \textsc{Grokfast-MA} & \textsc{Grokfast-EMA} & \textsc{NeuralGrok} & acc. rate (v.s. standard / MA / EMA) \\ 
 \midrule
 $a+b \ (\text{mod }97)$ & $1650$ & $1780$ &$1820$ &$\textbf{900}$ & $1.83\times$ / $1.98\times$ / $2.02\times$\\ 
 $a-b \ (\text{mod }97)$ & $4330$ & $2990$ & $\textbf{1340}$& $1467$ &$2.95\times$ / $2.04\times$/ $0.91\times$ \\ 
  $a\cdot b \ (\text{mod }97)$ &  $1280$ & $1150$ & $1400$ & $\textbf{918}$& $1.39\times$ / $1.25\times$ / $1.53\times$ \\ 
   $a^2-b \ (\text{mod }97)$ & $4820$ & $3830$ & $2730$ & $\textbf{1845}$ & $2.61\times$ / $2.08\times$ / $1.48\times$ \\ 
    $ac+bd-e \ (\text{mod }7)$ & - & $8853$ & - & $\textbf{1896}$ & - / $4.67\times$ / -\\ 
 \bottomrule
 \end{tabular}
\end{adjustbox}
\end{table}

\subsection{Effect of Gradient Rescaling}
According to \autoref{equ:neuralgrok}, the transformation of \textit{neural-amplifier} on the original gradient $\vg$ can be decomposed into two consecutive mechanisms: Firstly, it performs a rotation with a norm-1 vector $\vp\in\Delta^{|g|}$; then a magnitude rescaling is applied, which compresses or scales the gradient to a constant magnitude $c$. 
To investigate the effect of the gradient magnitude, we conduct comprehensive ablations on the hyperparameter $c$ under standard training and inside the \textsc{NeuralGrok} pipeline.
When applying standard training with weight-decay regularization, we apply a gradient normalization by: $ \vg'=c\cdot \frac{\vg}{\|\vg\|_2}$, which modifies the magnitude of the gradient without changing the direction. 
Note that gradient rescaling is not equivalent to applying various learning rates, as the learning rate does not render a constant gradient magnitude but can be seen as a constant amplification at every training step.
\begin{figure}[!ht]
      \centering
	   \subfloat[$c=0.01$.]
		{\includegraphics[width=0.24\textwidth]{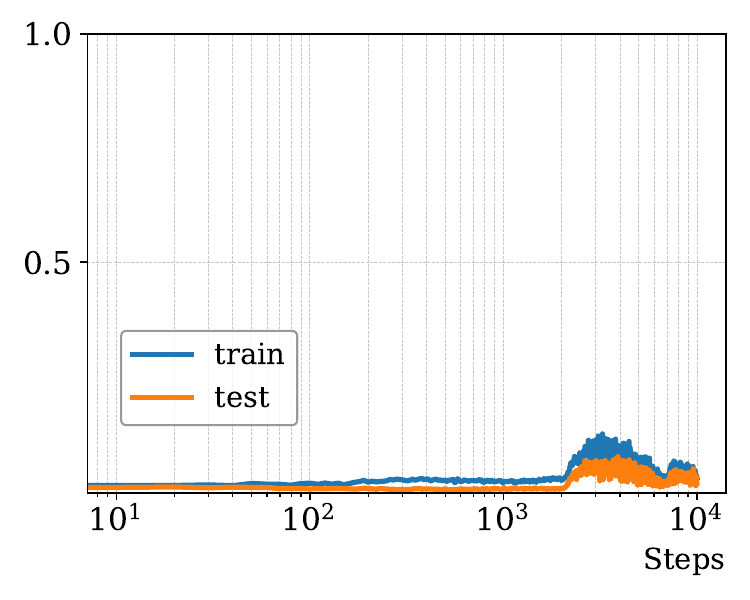}}
        \hfill
	   \subfloat[$c=0.5$.]
		{\includegraphics[width=0.24\textwidth]{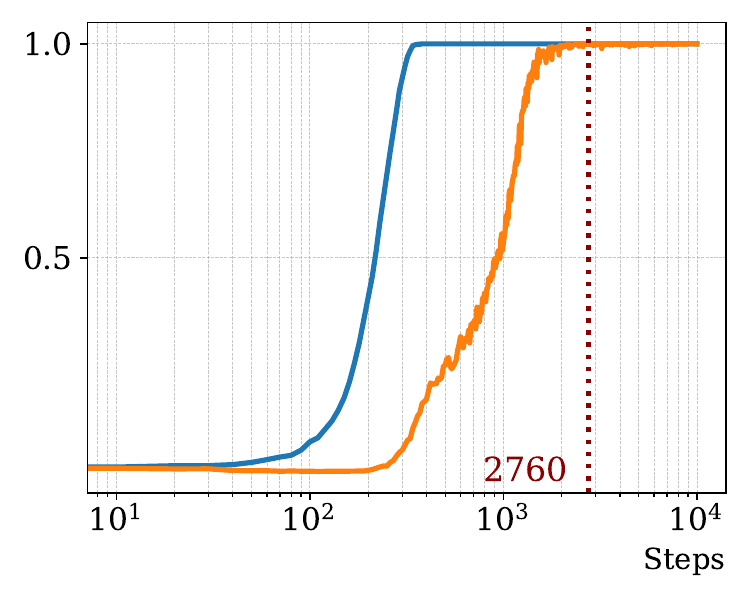}}
        \hfill
      \subfloat[$c=1.0$.]
		{\includegraphics[width=0.24\textwidth]{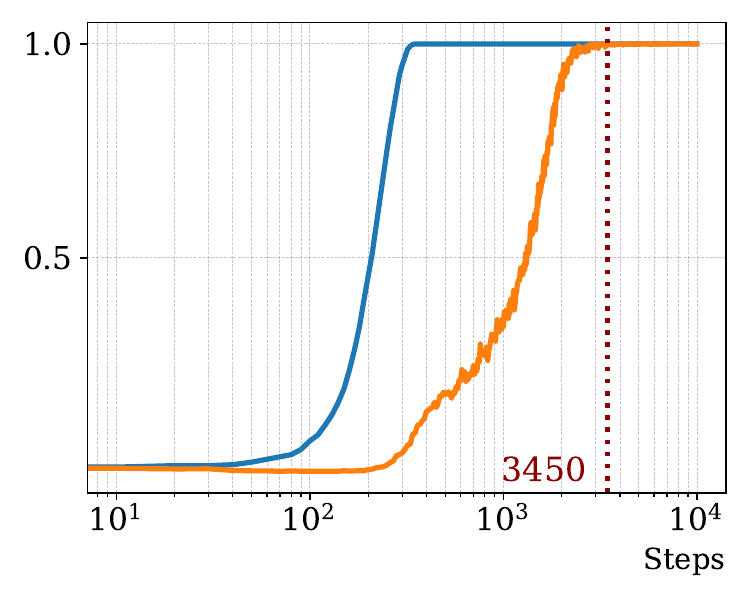}}
        \hfill
      \subfloat[$c=2.0$.]
		{\includegraphics[width=0.24\textwidth]{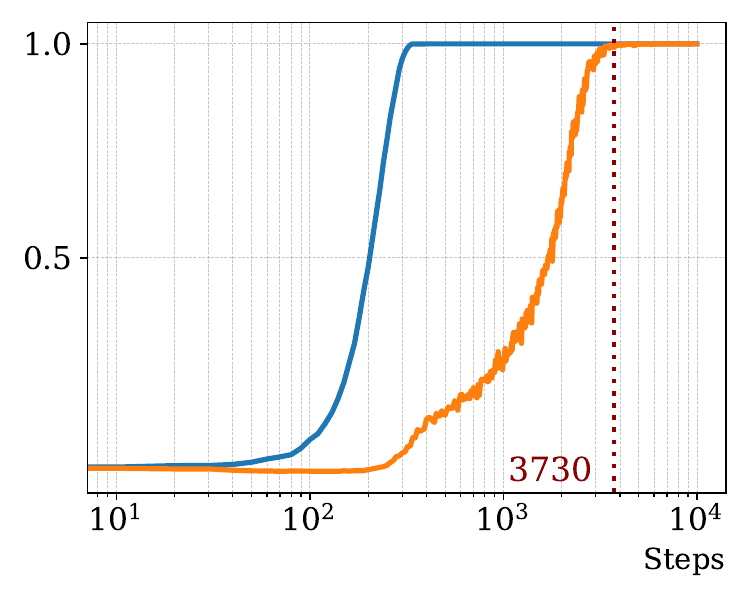}}
	\caption{\textbf{Standard training with various gradient rescaling coefficient $c$ on \texttt{(a+b) mod 97} task.} With $c$=$0.5,1.0,2.0$, the training is effectively stabilized compared to \autoref{fig:all_tasks_std} with unchanged magnitude.}
	\label{fig:std_w_gradnorm}
\end{figure}

\begin{figure}[!ht]
\centering
\begin{adjustbox}{max width=0.9\textwidth}
\centering
	   \subfloat[\texttt{(a+b) mod 97}.]
		{\includegraphics[width=0.32\textwidth]{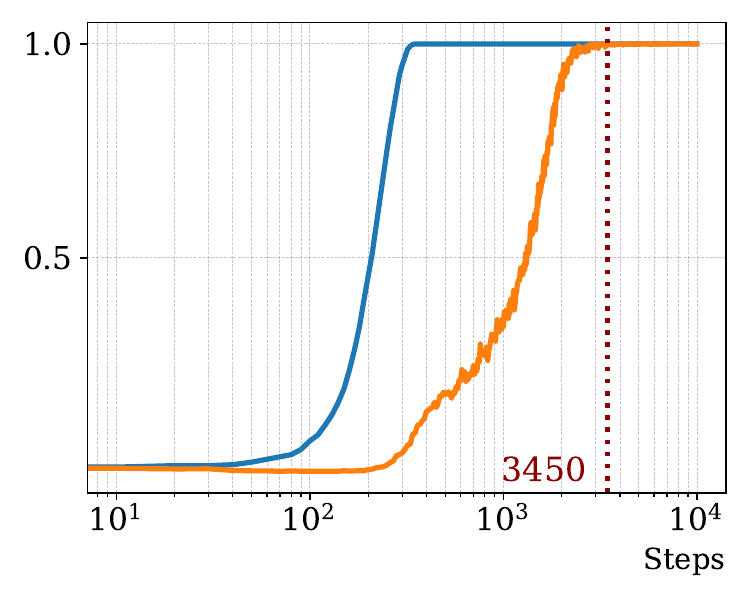}}
        \hfill
	   \subfloat[\texttt{(a$\times$a-b) mod 97}.]
		{\includegraphics[width=0.32\textwidth]{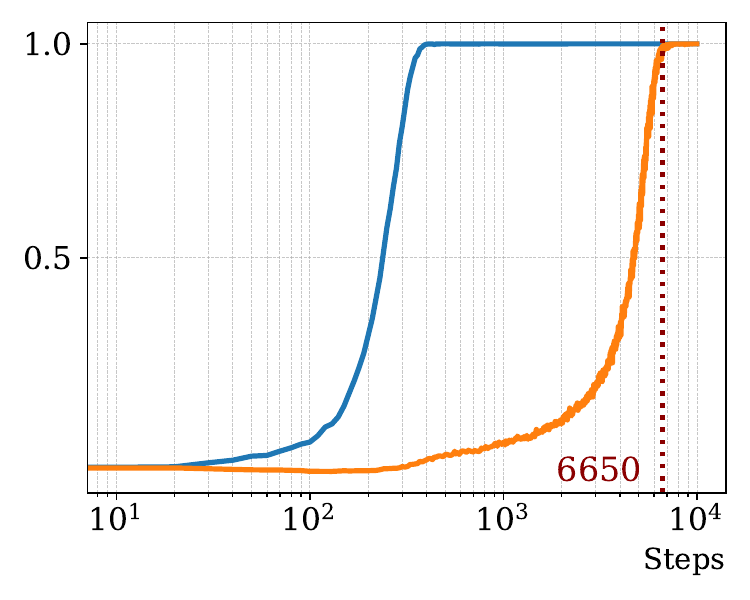}}
        \hfill
      \subfloat[\texttt{(a$\times$c+b$\times$d-e) mod 7}.]
		{\includegraphics[width=0.32\textwidth]{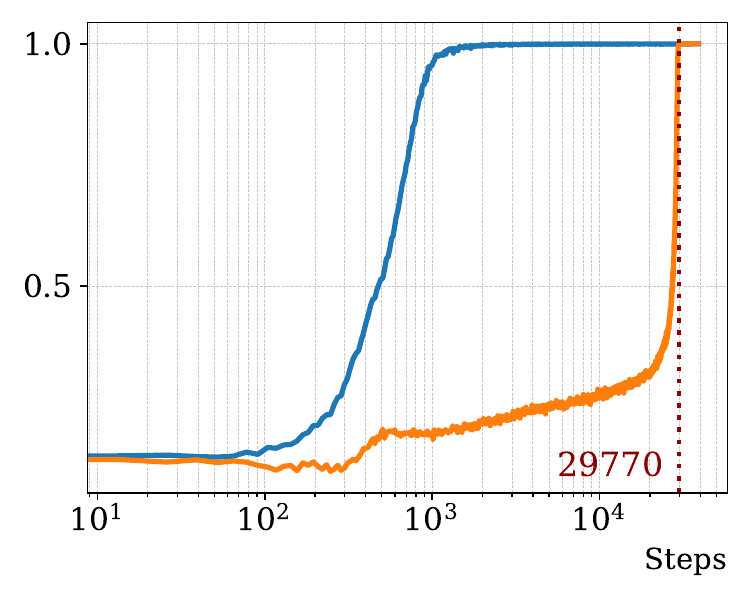}}
\end{adjustbox}
    \caption{\textbf{Standard training with standard gradient normalization $c$=$1.0$.} The gradient normalization enables the generalization on the challenging task \texttt{(a$\times$c+b$\times$d-e) mod 7}, which is failed in \autoref{fig:all_tasks_std}, when gradient normalization is not applied.}
\label{fig:std_w_gradnorm_3tasks}
\end{figure}

\textbf{Gradient Rescaling as a Better Regularization than Weight-decay.}
By simply applying gradient normalization with standard training, not only the training dynamics are stabilized, but the generalization is accelerated, especially on challenging tasks.
According to \autoref{fig:std_w_gradnorm}, with gradient rescaled to $c$=$0.5,1.0,2.0$, the accuracy scores on both training and test sets are greatly stabilized without significant spikes. Notably, with $c$=$0.5$, the generalization on the test set is mostly accelerated compared to a larger gradient scale ($c$=$1.0,2.0$). However, the training collapses when $c$ decreased to $0.01$, where the learning is significantly slowed down due to small gradient updates. When apply a standard gradient normalization $c$=$1.0$ on various tasks (\autoref{fig:std_w_gradnorm_3tasks}), the transformer model is able to learn the challenging task \texttt{(a$\times$c+b$\times$d-e) mod 7}, which is failed in \autoref{fig:all_tasks_std}, with the original unchanged gradient magnitude. It indicates that gradient rescaling can be a more effective regularization than conventionally used weight-decay on arithmetic task learning.

\paragraph{\textsc{NeuralGrok} is robust with various rescaling coefficients.}
While gradient rescaling acts as a crucial factor when applying standard training on transformer models, \textsc{NeuralGrok} exhibits a robust generalization performance with various values of rescaling coefficient $c$. 
We present the training/test accuracies with $c$ ranging from $0.2$ to $2.0$. 
The transformer model consistently achieve a perfect test accuracy with similar speed($1.3k steps$) with $c$ ranging from $0.2$ to $1.0$. While applying a larger gradient magnitude $c$=$2.0$ could lead to a delayed generalization, reaching the perfect test accuracy at $\sim2320$ steps.
It suggests that the \textit{neural-amplifier} can effectively adapt to different gradient magnitudes in the outer-loop update, which further demonstrates the robustness and learning capacity of \textsc{NeuralGrok}. 
\begin{figure}[!ht]
\centering
\begin{adjustbox}{max width=0.99\textwidth}
\centering
	   \subfloat[$c=0.2$.]
		{\includegraphics[width=0.32\textwidth]{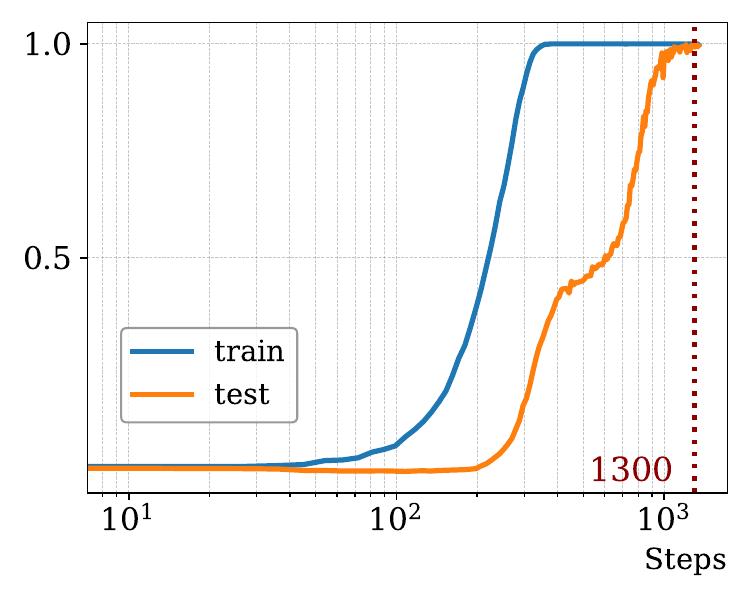}}
        \hfill
	\subfloat[$c=0.5$.]
		{\includegraphics[width=0.32\textwidth]{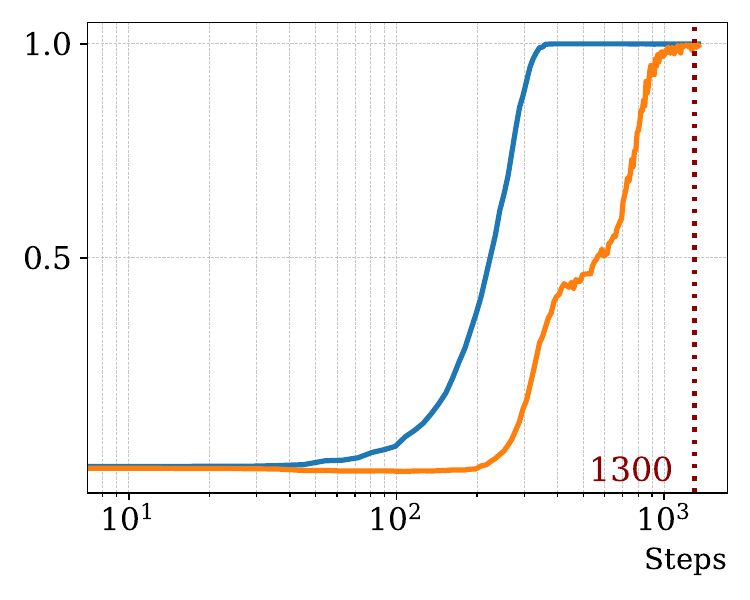}}
        \hfill
        \subfloat[$c=1.0$.]
		{\includegraphics[width=0.32\textwidth]{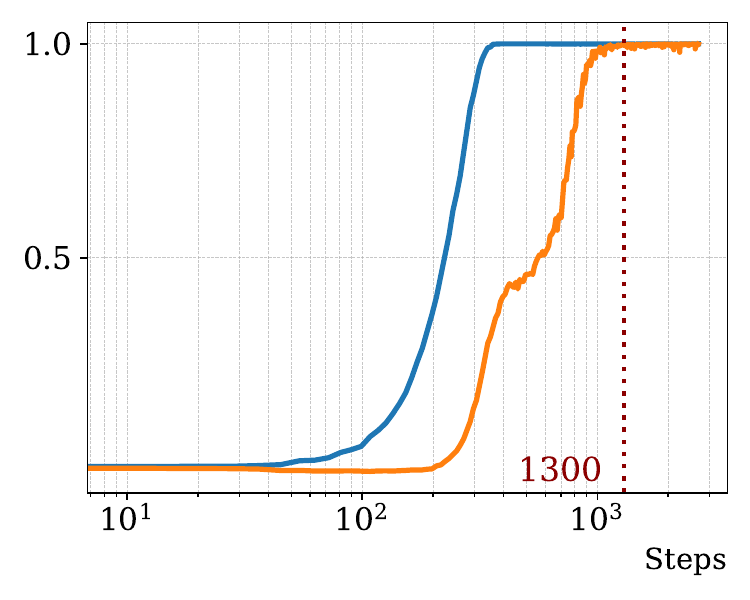}}
        \hfill
      \subfloat[$c=2.0$.]
		{\includegraphics[width=0.32\textwidth]{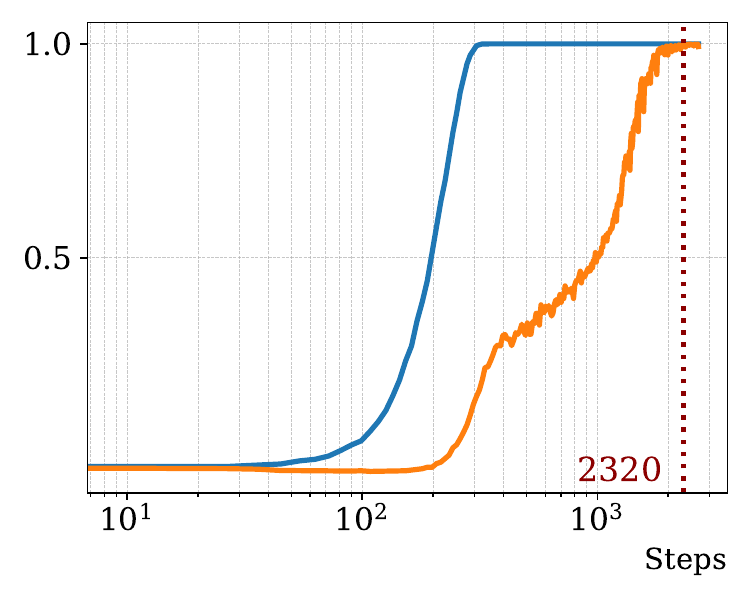}}
\end{adjustbox}
	\caption{\textsc{NeuralGrok} with various gradient rescaling coefficient $c$ on the \texttt{(a+b) mod 97} task. The model can achieve a perfect test accuracy in similar speed ($1.3k steps$) with $c$ ranging from $0.2$ to $1.0$. While applying a larger gradient magnitude $c$=$2.0$ could lead to a delayed generalization.}
	\label{fig:coefficient-neuralgrok}
\end{figure}

\section{Interpret Grokking with Weight and Gradient Complexity}
Prior studies on the grokking phenomenon have proposed valuable theoretical and empirical insights on the memorization-to-generalization phase transition. \citet{liu2023omnigrokgrokkingalgorithmicdata} proposed that the model achieves generalization when the model weights are optimized into a \textit{Godilocks zone}, which correlates with the decrease of the Euclidean norm of model weights. 
However, recent research \citep{demoss2024complexitydynamicsgrokking, golechha2024progressmeasuresgrokkingrealworld} argues that the dynamics of the weight norm cannot well explain the phase transition under the grokking phenomenon. 
Alternatively, \citet{golechha2024progressmeasuresgrokkingrealworld} proposed to apply the \textbf{A}bsolute \textbf{W}eight \textbf{E}ntropy (AWE) as an assessment of a model's complexity:
\begin{align}\label{equ:awe}
    H(\mathcal{W}) = -\sum_{w_{i}\in \mathcal{W}} |w_{i}|\ln{|w_{i}|},
\end{align}
where $\mathcal{W}$ denotes a given weight vector or matrix. Following the AWE metric, we further measure the \textbf{A}bsolute \textbf{G}radient \textbf{E}ntropy (AGE) score during training, which reflects the instantaneously acquired complexity at the current optimization step:
\begin{align}\label{equ:age}
    H(\mathcal{G}) = -\sum_{g_{i}\in \mathcal{G}} |g_{i}|\ln{|g_{i}|},
\end{align}
where $\mathcal{G}$ denotes a given gradient vector or matrix. We then measure the evolution of both AWE and AGE scores throughout the training runs to show how they correlate with the memorization and generalization progress.
\begin{figure}[!ht]
     \centering
     \begin{subfigure}[b]{0.85\textwidth}
         \centering
         \includegraphics[width=\textwidth]{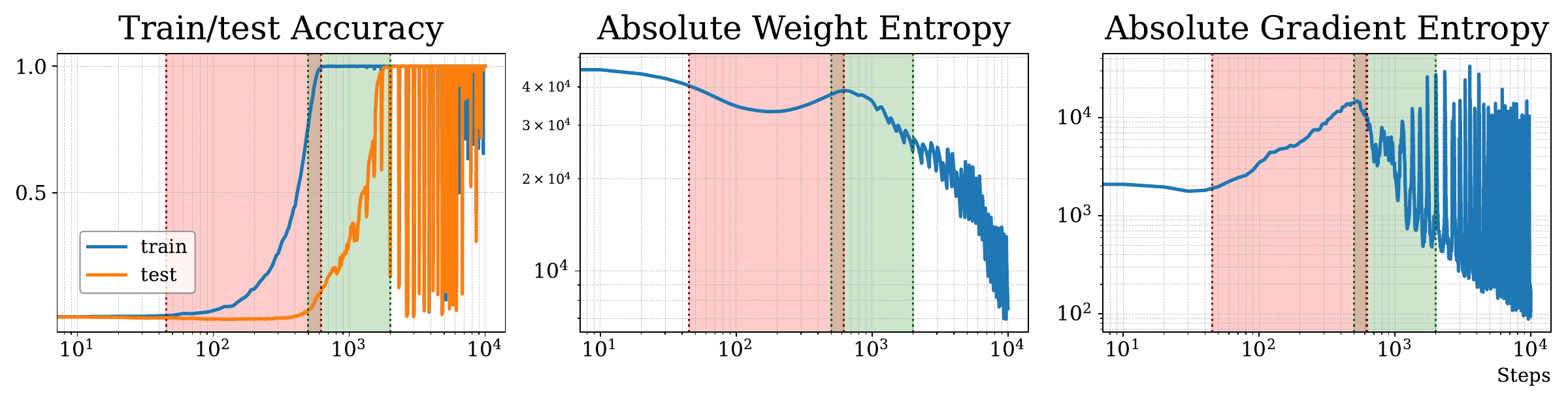}
         \vspace{-2em}
         \caption{\small Standard Training.}
         \label{fig:std-entropy}
     \end{subfigure}
     \hfill
     \begin{subfigure}[b]{0.85\textwidth}
         \centering
         \includegraphics[width=\textwidth]{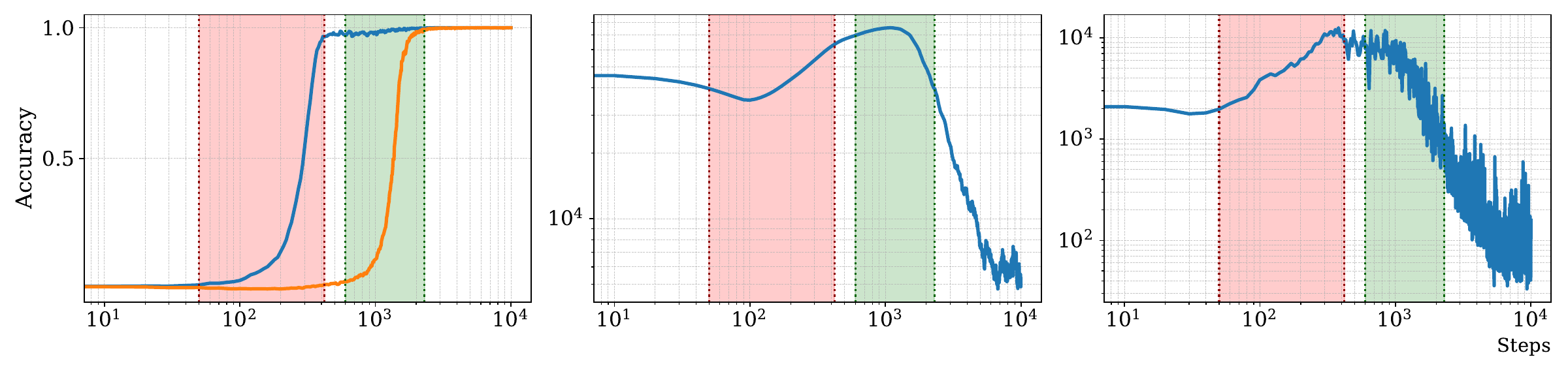}
         \vspace{-2em}
         \caption{\small \textsc{GrokFast}-MA.}
         \label{fig:ma-entropy}
     \end{subfigure}
     \hfill
     \begin{subfigure}[b]{0.85\textwidth}
         \centering
         \includegraphics[width=\textwidth]{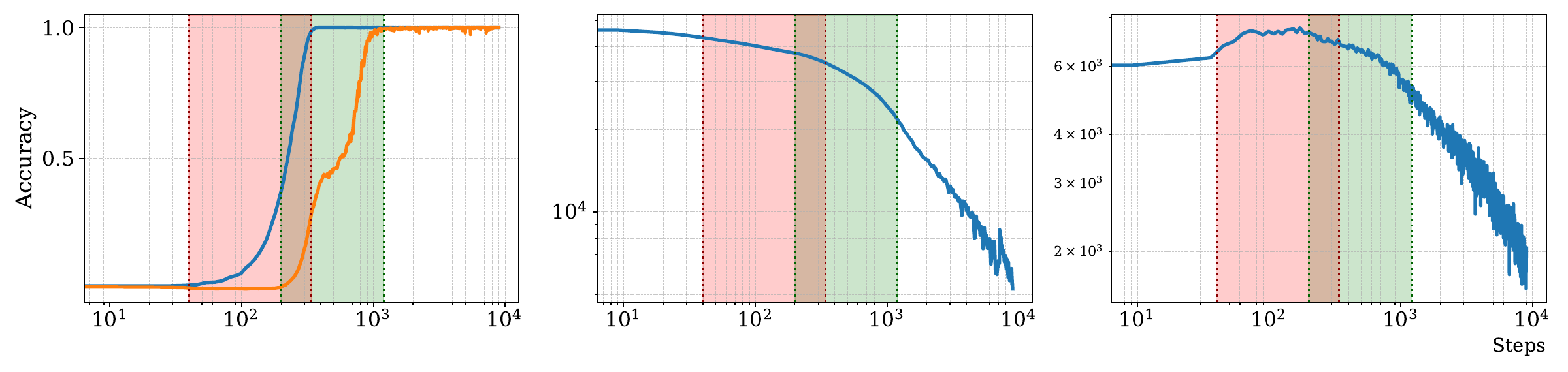}
         \vspace{-2em}
         \caption{\small \textsc{NeuralGrok}.}
         \label{fig:ng-entropy}
     \end{subfigure}
        \caption{\textbf{Model complexity measured in entropy on task \texttt{(a+b) mod 97}.} The transition windows for \textcolor{DarkRed}{Memorization} and \textcolor{DarkGreen}{Generalization} phases are marked in \textcolor{DarkRed}{red} and \textcolor{DarkGreen}{green}.}
        \label{fig:all-entropy}
\end{figure}
\paragraph{Absolute Gradient Entropy as an effective indicator of phase transitions.} As shown in \autoref{fig:all-entropy}, we train the transformer models on task \texttt{(a+b) mod 97} while reporting the AWE and AGE scores along with the training/test accuracy curves. 
In each experiment, we mark the transition windows for the Memorization and Generalization phases, respectively, in red and green colors. Across all three experiments, the evolution of AGE scores demonstrates a remarkable correspondence to the phase transitions: 
\textbf{In the \textit{memorization} phase}, where the training accuracy increases from zero to a perfect level, \textbf{the AGE score increases accordingly, suggesting the model is fitting onto a sophisticated feature space}; while \textbf{in the \textit{generalization} phase}, where the model starts adapting to the heldout set with an increasing test accuracy, \textbf{the AGE score decreases, indicating the model gradually compresses the memorized features into a generalizable pattern}.

\begin{wrapfigure}{r}{0.5\textwidth}
\centering
    \includegraphics[width=\linewidth]{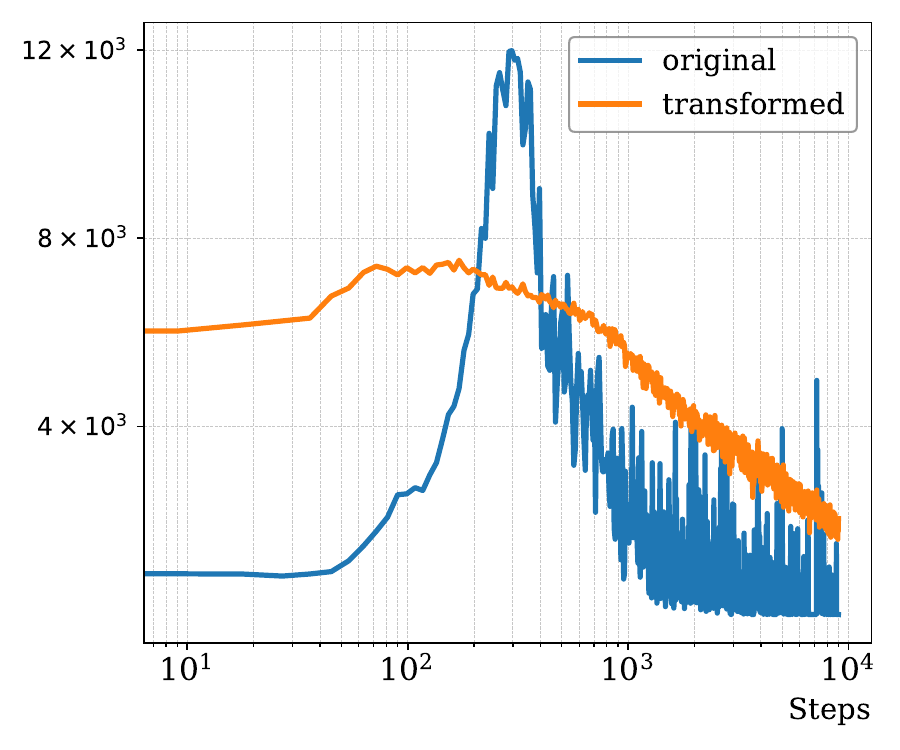}
    \caption{\textbf{Absolute Gradient Entropy before and after gradient transformation by the \textit{neural-amplifier}.} }
    \label{fig:neuralgrok-entropy-compare}
\end{wrapfigure}

\paragraph{\textsc{NeuralGrok} accelerate generalization by reducing model complexity.} 
Compared to Standard training (\autoref{fig:std-entropy}) and \textsc{GrokFast}-MA (\autoref{fig:ma-entropy}), the model trained with \textsc{NeuralGrok} exhibits lower scores of AWE and AGE, suggesting a lower model complexity and better generalizability. 
In \autoref{fig:neuralgrok-entropy-compare}, we present the AGE scores of the original gradients, and the transformed gradients after the transformation by the \textit{neural-amplifier}. The original gradients before transformation exhibit a significant complexity spike around $5\times 10^2$, while the transformed gradients evolving smoothly.

\section{Discussion and Limitations}
\paragraph{Is Weight-decay always a good regularization?} 
While previous studies claim that weight-decay is the crucial factor to enable generalization under the grokking phenomenon, we observe that adding weight-decay may not help but impede the learning on the challenging arithmetic tasks. 
We investigate various combination of regularization techniques on task \texttt{$(a\times c+b\times d-e)$ mod 7} and present the results in \autoref{fig:std-wd-rescale-ablation-ac+bd-e}.
We find that applying the standard gradient normalization alone can effectively stabilize the training and slightly accelerate the generalization under the grokking context. 
Conversely, only apply the weight-decay causes a catastrophic collapse in \autoref{fig:std-wd-rescale-ablation-ac+bd-e} (c), where the model stops learning from the task without generalization nor memorization happening.
In practice, we recommend applying a small value of weight decay combined with standard gradient normalization to achieve the best performance on challenging arithmetic tasks.

\paragraph{Surprisingly low transferability of the gradient transformations.} 
While \textsc{NeuralGrok} accelerates grokking within individual arithmetic tasks, we find that the learned gradient transformations exhibit limited transferability even across operations leveraging similar correlations between variables and operators (e.g., $+$ vs. $-$). 

This suggests that the \textit{neural-amplifier} adapts to highly task-specific gradient patterns, such as suppressing noise in cyclic modular operations or amplifying critical features to disentangle composite equations. 
For instance, transformations optimized for modular addition ($+$) may fail to generalize to subtraction ($-$) or mixed-operation tasks (\autoref{fig:(a-b)-transfer}), where gradient dynamics not only cares about correlations between variables and operators, but also involve nuance reasoning mechanisms. 
This specialization might stem from the interplay between the bilevel optimization framework and the narrow validation objectives ($\mathcal{D}_{outer}$), which force the amplifier to local task geometries rather than global arithmetic principles. Future work could investigate cross-task meta-learning or shared amplification modules to disentangle universal arithmetic patterns from task-specific adaptations.

\paragraph{Limited datasets and task settings.} We currently only conduct experiments on the synthetic arithmetic tasks, which provides us a perfect testbed for with controllable setting where we can design experiments to decouple the factor which impacts grokking from real-world noises or dataset biases.
With the promising performance on the arithmetic tasks, we expect to extend the bilevel formulation and the insight of learnable neural gradient amplifier to more complex application domains, e.g. LLM training, etc. We extend it as future work.

\section{Related Work}

\paragraph{Empirical Observation of Grokking.} The phenomenon of grokking—delayed generalization after prolonged overfitting—was first empirically observed by \citet{power2022grokking} in transformer models trained on arithmetic tasks. This discovery spurred a wave of research into understanding the dynamics of memorization and generalization in over-parameterized networks. Subsequent studies explored grokking across diverse tasks \citep{power2022grokking, liu2023omnigrokgrokkingalgorithmicdata, lee2024grokfastacceleratedgrokkingamplifying}. \citet{liu2022understandinggrokkingeffectivetheory} and \citet{kumar2024grokkingtransitionlazyrich} further investigated grokking through the lens of representation learning, identifying phase transitions in model behavior during training. Notably, \citet{lee2024grokfastacceleratedgrokkingamplifying} demonstrated that manipulating gradient signals, such as amplifying low-frequency components via a low-pass filter, could significantly accelerate generalization. Empirical analyses by \citet{googlepair2023grokking} and \citet{demoss2024complexitydynamicsgrokking} revealed that models transition from dense, high-magnitude weight configurations during memorization to sparse, simpler structures during generalization, a pattern corroborated by metrics like Absolute Weight Entropy (AWE) \citep{golechha2024progressmeasuresgrokkingrealworld}. These observations highlight the critical role of training dynamics and regularization in shaping grokking behavior.

\paragraph{Theoretical Understanding of Grokking.} Theoretical efforts to explain grokking have focused on optimization dynamics, model complexity, and implicit regularization. \citet{davies2023unifyinggrokkingdoubledescent} unified grokking with the double-descent phenomenon, attributing delayed generalization to the interplay between model capacity and data complexity. \citet{thilak2022slingshotmechanismempiricalstudy} linked grokking to adaptive optimization strategies, showing that gradient noise and sharp minima influence generalization timing. \citet{krogh1991simple} and \citet{xie2024overlookedpitfallsweightdecay} emphasized the dual role of weight decay: while it promotes generalization by controlling model complexity, excessive decay can destabilize training and impede convergence. \citet{hardt2016trainfastergeneralizebetter} and \citet{li2020generalizationerrorboundsnoisy} connected gradient norm stability to generalization, suggesting that sharp minima—associated with large gradient norms—correlate with poor extrapolation. 

\section{Conclusion}
In this paper, we propose a bilevel optimization framework \textsc{NeuralGrok} as a novel approach that learns an optimal gradient transformation to accelerate the generalization of transformers in arithmetic tasks. 
Through extensive experiments on arithmetic tasks, we demonstrate that \textsc{NeuralGrok} effectively facilitate the generalization while also stabilising the training dynamics. 
We further proposed the Absolute Gradient Entropy metric as a measurement of the learning complexity at each optimization steps. We discover that Absolute Gradient Entropy consistently correlates with the phase transitions under grokking phenomenon, including memorization and generalization stages.

\bibliography{arxiv2025}
\bibliographystyle{colm2025_conference}

\appendix

\section{Arithmetic Datasets}
\label{appdix:algorithm-dataset}
We apply the similar arithmetic dataset construction strategy in \citet{power2022grokking}. However, we do not just assign a single operator $\langle op \rangle$ to represent complex mathematical expressions with more than one operators. For example, in \citet{power2022grokking}, for mathematical expression $x^2+xy+y^2$, they only use one single operator $\langle op \rangle$: $\circ$ to express $x \circ y = x^2+xy+y^2$, and construct the dataset of equations of the form $\langle x\rangle \langle op\rangle \langle y \rangle\langle =\rangle \langle x\circ y\rangle$, where $\langle a\rangle$ stands for the token corresponding to element $a$.

In our experiments, we assign different operators (e.g. $+,-,\times$) to different tokens: $\langle op_1\rangle, \langle op_2\rangle,...$. Moreover, we do not only limit in binary operators with only two variables $x,y$, but also extend to more variables to increase datasets difficulty. Formally speaking, suppose one mathematical expression involves with $n$ variables $v_1, v_2, ..., v_n$ and $m$ different mathematical operators $op_1, op_2,...,op_m$, we construct the dataset of equations as follows:
\begin{align*}
    \langle v_1\rangle\langle v_2\rangle...\langle v_n\rangle \langle op_1\rangle \langle op_2\rangle...\langle op_m\rangle \langle =\rangle \langle ans \rangle
\end{align*}
where $ans$ denotes the answers of the mathematical equations. All arithmetic tasks are under modular arithmetic, with a prime number $p$. Taking $a+b\ (\text{mod }97)$ for example, the dataset is constructed in the following format:
\begin{align*}
    \langle a\rangle\langle b\rangle\langle+\rangle\langle =\rangle\langle ans\rangle
\end{align*}
Since each input variable $v_i$ can be chosen between $0$ and $p-1$, the total amount of one task with $n$ variables would approximate $p^n$.

\newpage
\section{Neural-Amplifier Implementation} \label{appd:mlp}
\subsection{Detailed Architecture}
The \textit{neural-amplifier} $G(\varphi)$ contains simple MLPs to process the original gradients $\vg$. In our main experiments, we set the hidden dimension $d=32$. We use ReLU \citep{agarap2019deeplearningusingrectified} as the activation function and normalize the transformed gradient after the Softmax operation to get the final modified gradient $\textbf{g}'$. In the main experiments, we set $c=1$ in \autoref{equ:neuralgrok}. We provide the PyTorch implementation as follows:

\lstset{frame=tb,
  language=Python,
  aboveskip=3mm,
  belowskip=3mm,
  showstringspaces=false,
  columns=flexible,
  basicstyle={\small\ttfamily},
  breaklines=true,
  breakatwhitespace=true,
  tabsize=3,
  backgroundcolor=\color{backcolour},   
    commentstyle=\color{codegreen},
    keywordstyle=\color{magenta},
    numberstyle=\tiny\color{codegray},
    stringstyle=\color{codepurple},
}
\begin{figure*}[!h]
    \centering
    \begin{lstlisting}
class NeuralGrok(nn.Module):
    def __init__(self, hidden_dim=32, n_layers=2, alpha=16):
        super(NeuralGrok,self).__init__()

        self.alpha = alpha

        hidden_dim_alpha = int(self.alpha * hidden_dim)

        layers = []

        layers.append(nn.Linear(1, hidden_dim_alpha))
        layers.append(nn.ReLU())
       
        for i in range(n_layers-1):
            if i == n_layers-2:
                layers.append(nn.Linear(hidden_dim_alpha, 1))
            else:
                layers.append(nn.Linear(hidden_dim_alpha, hidden_dim_alpha))
                layers.append(nn.ReLU())

        self.mlp = nn.Sequential(*layers)
        self.softmax = nn.Softmax(dim=0)
        
    def forward(self, grad):
        mlp1 = self.mlp(grad)
        p = self.softmax(mlp1)
        x = p * grad / torch.norm(p * grad)
        return x
    
    \end{lstlisting}
     \caption{Code for \textsc{NeuralGrok}}
     \label{code:neuralgrad}
\end{figure*}

\section{Supplement Results on Baselines}\label{appd:baseline}
\subsection{Standard Training with Various \textit{weight-decay}}
\begin{figure}[!ht]
  \centering
  \subfloat[$a+b\  (\text{mod }97)$.]{\includegraphics[width=0.24\textwidth]{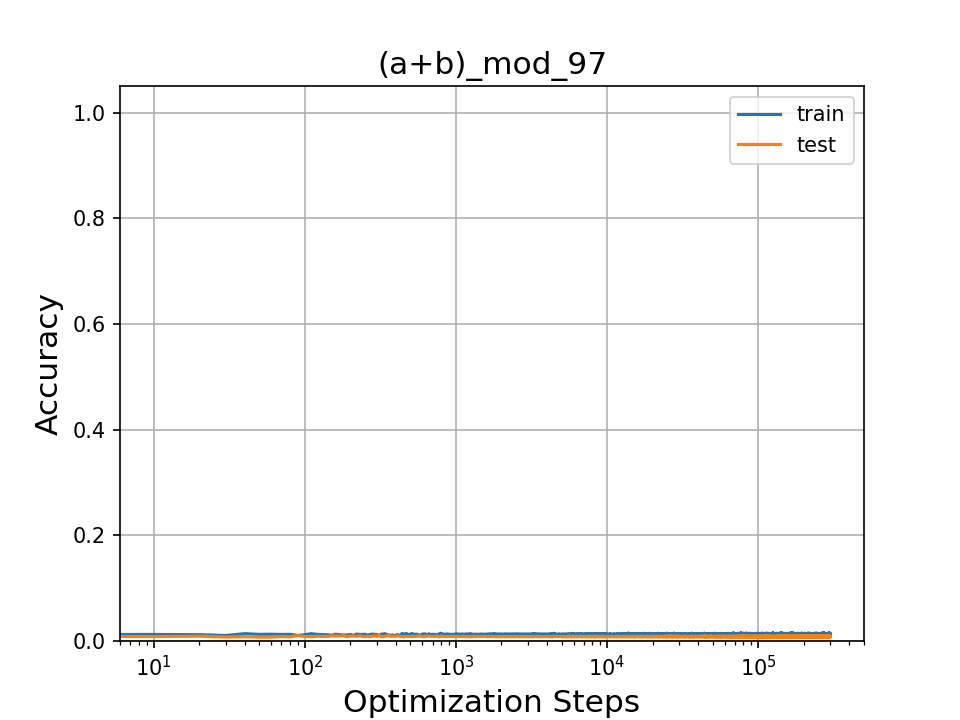}}
  \hfill
  \subfloat[$a-b\  (\text{mod }97)$.]{\includegraphics[width=0.24\textwidth]{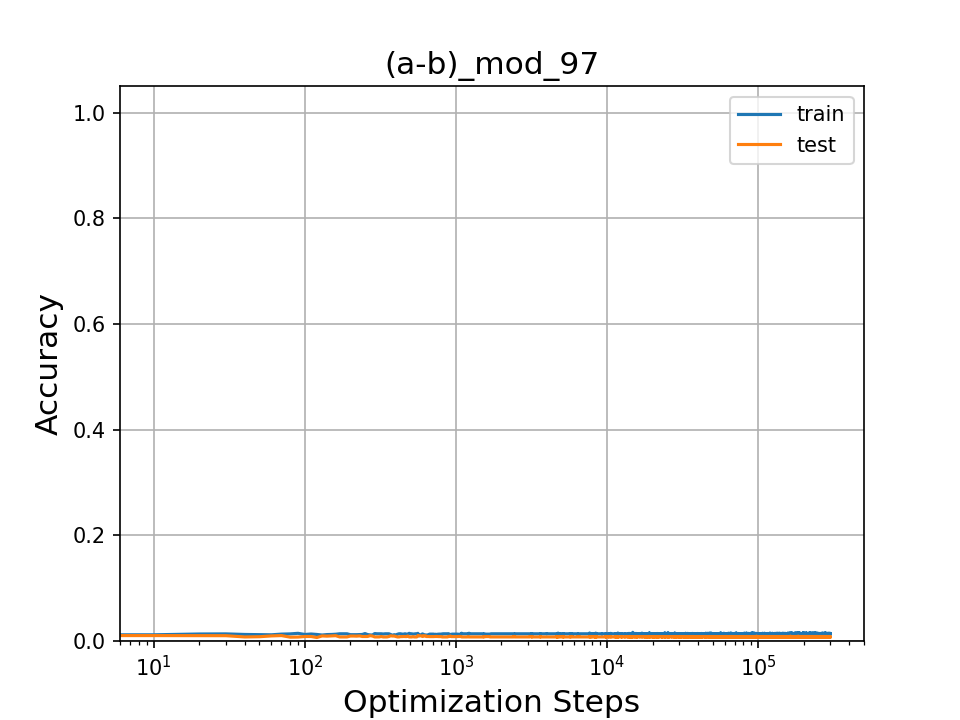}}
  \hfill
  \subfloat[$a^2- b\ (\text{mod }97)$.]{\includegraphics[width=0.24\textwidth]{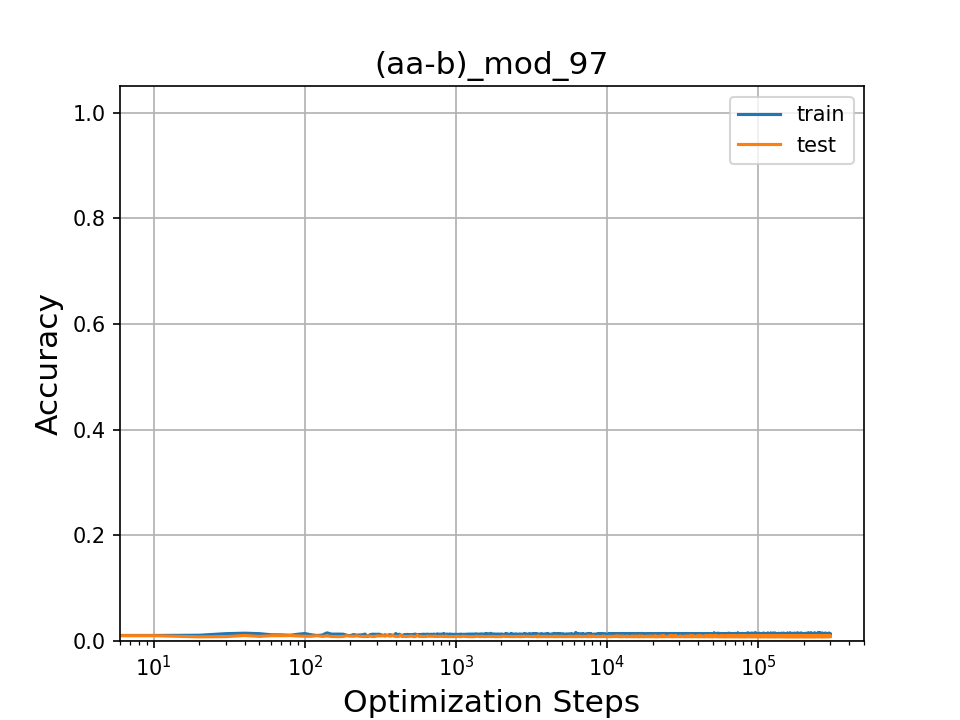}}
  \hfill
  \subfloat[$ac+bd-e\ (\text{mod }7)$.]{\includegraphics[width=0.24\textwidth]{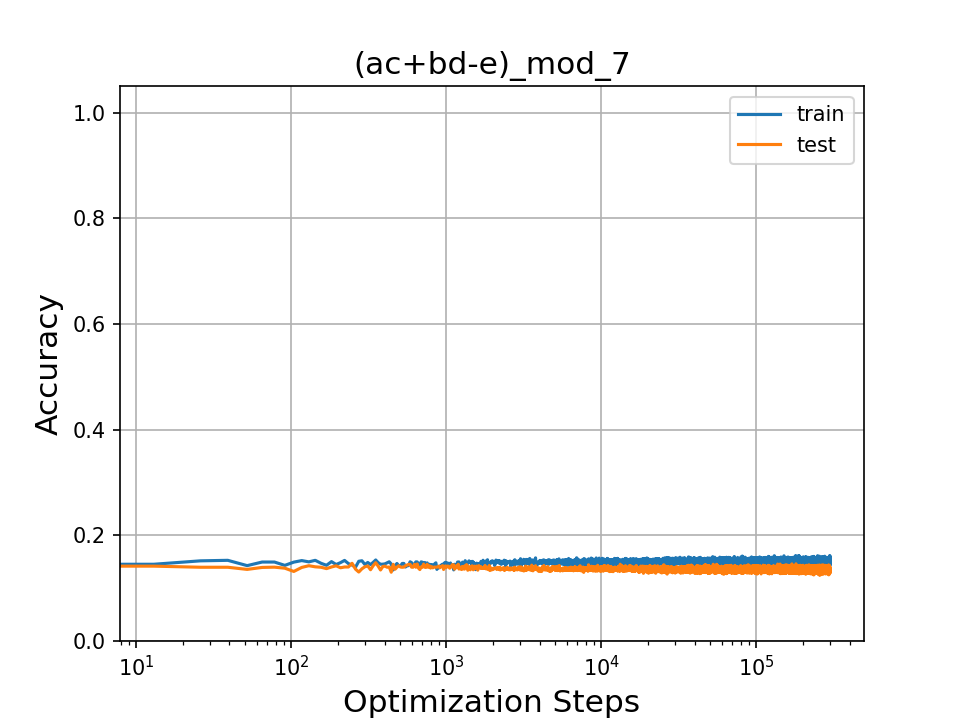}}
  \caption{Standard training with \textit{weight-decay} $1e^{-2}$. The model fails to memorize and generalize in all four experiments.}
  \label{appd:fig:std-training-wd=0.01}
\end{figure}

We try two different values (i.e., $1e^{-2}, 1e^{-3}$) of \textit{weight-decay} to observe the learning pattern of the model in the standard training. However, we find that when \textit{weight-decay} is set to be larger (i.e., $1e^{-2}$), the model fails to memorize and generalize, which is the reason why we choose a smaller value $1e^{-3}$ for standard training as the baseline. The results are shown in \autoref{appd:fig:std-training-wd=0.01}. 

\newpage
\subsection{\textsc{GrokFast-MA} with Various \textit{weight-decay}}
\begin{figure}[!ht]
      \centering
	   \subfloat[$a+ b\ (\text{mod }97)$ with $wd=1e^{-2}$.]
		{\includegraphics[width=0.45\textwidth]{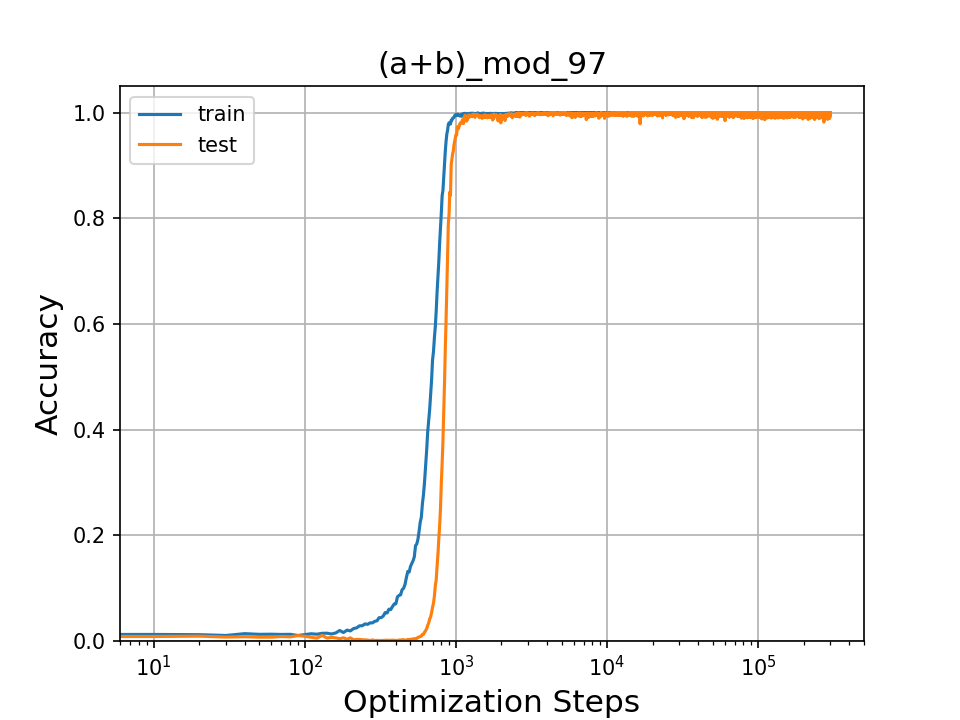}}
        \hfill
	   \subfloat[$a+ b\ (\text{mod }97)$ with $wd=1e^{-3}$.]
		{\includegraphics[width=0.45\textwidth]{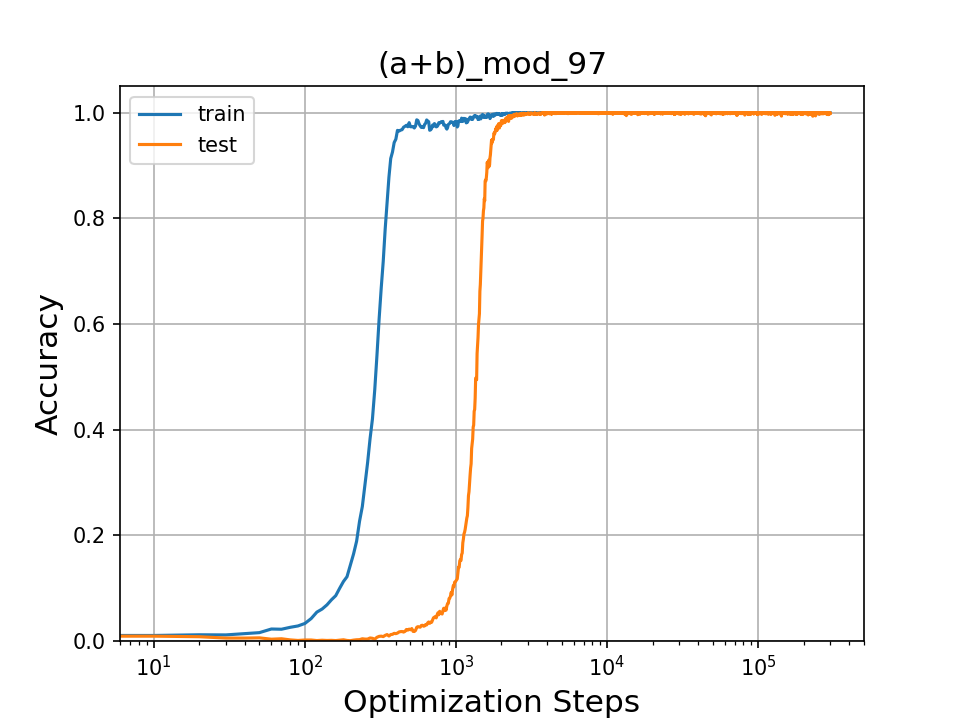}}
        \quad
      \subfloat[$ac+bd-e\ (\text{mod }7)$ with $wd=1e^{-2}$.]
		{\includegraphics[width=0.45\textwidth]{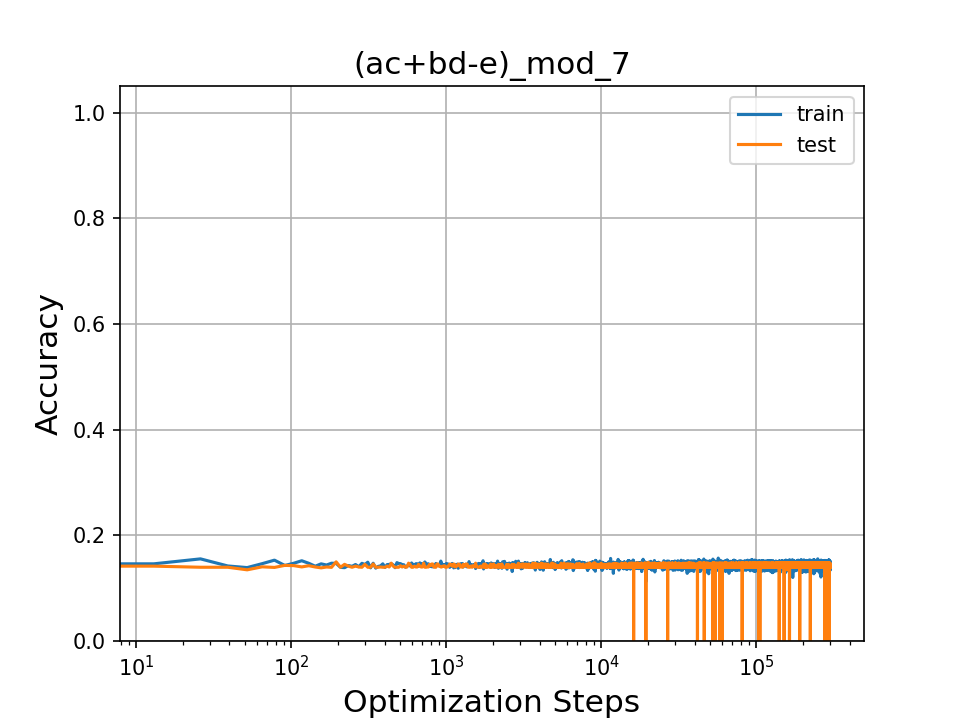}}
        \hfill
        \subfloat[$ac+bd-e\ (\text{mod }7)$ with $wd=1e^{-3}$.]
		{\includegraphics[width=0.45\textwidth]{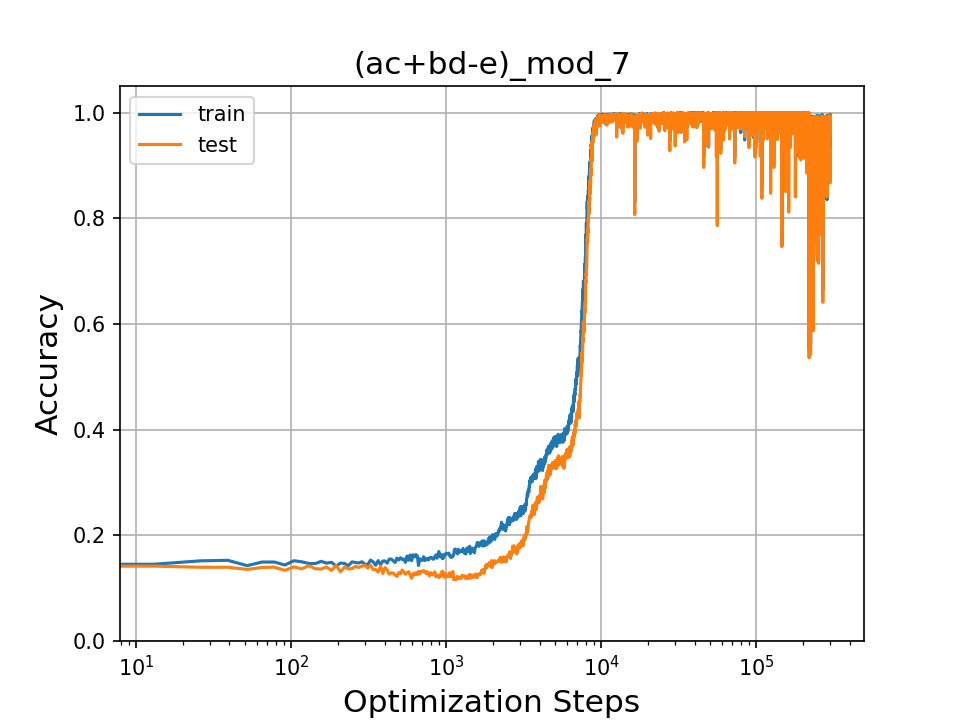}}
	\caption{\textsc{Grokfast-MA} training using different \textit{weight-decay} values.}
	\label{fig:ma-wd0.01-vs-wd0.001}
\end{figure}

We also compare different values (i.e., $1e^{-2}, 1e^{-3}$) of \textit{weight-decay} influence on \textsc{Grokfast-MA}. The results are shown in \autoref{fig:ma-wd0.01-vs-wd0.001}. 

From the figure, we can find that a larger \textit{weight-decay}, in some task (e.g., $a+b \ (\text{mod }97)$), can accelerate grokking better. However, in $ac+bd-e \ (\text{mod }7)$, the model cannot even learn in the same optimization steps. Therefore, we set \textit{weight-decay} as $1e^{-3}$ as the default setting in our main experiments. 
\newpage
\subsection{Experiments by \textsc{Grokfast-EMA}}
\begin{figure}[!ht]
  \centering
  \subfloat[\texttt{a+b (mod 97)}.]{\includegraphics[width=0.2\textwidth]{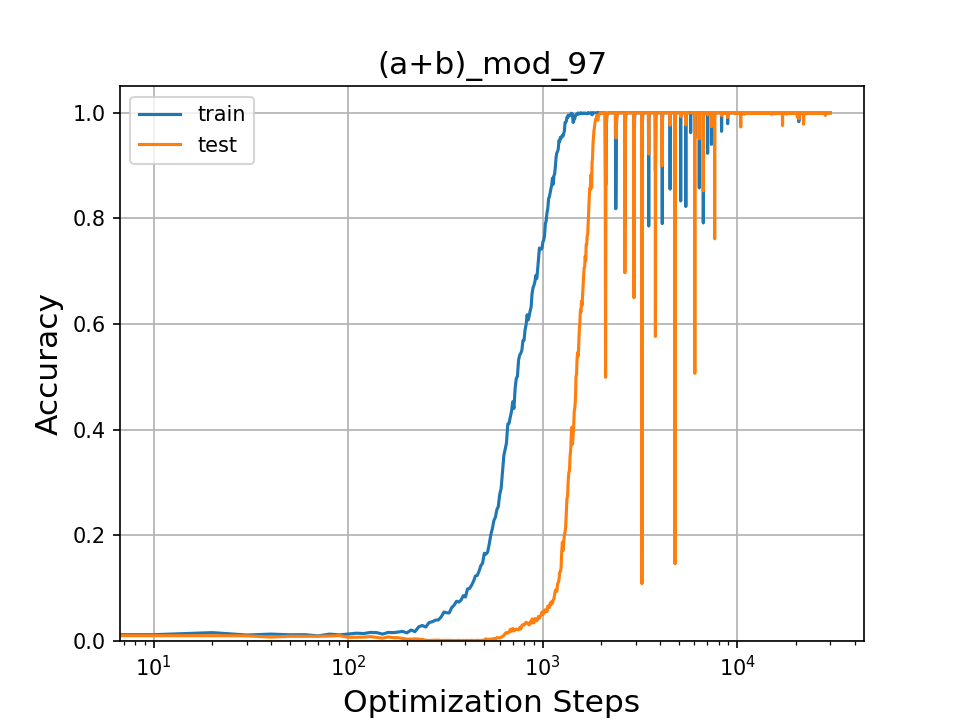}}
  \hfill
  \subfloat[\texttt{a-b (mod 97)}.]{\includegraphics[width=0.2\textwidth]{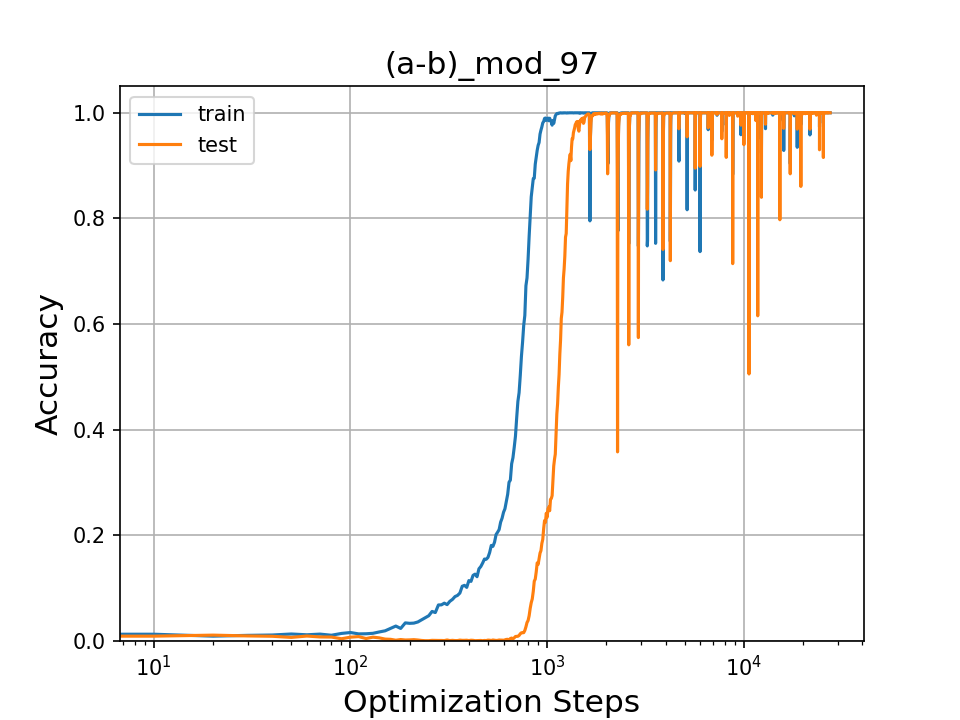}}
  \hfill
  \subfloat[\texttt{axb (mod 97)}]{\includegraphics[width=0.2\textwidth]{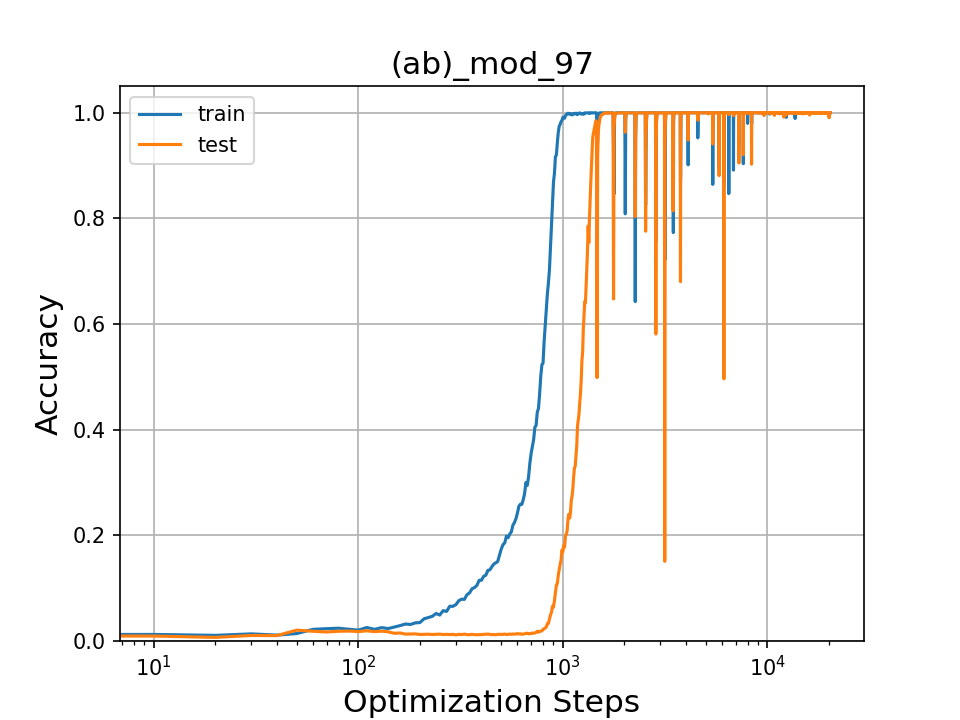}}
  \hfill
  \subfloat[\texttt{axa-b (mod 97)}.]{\includegraphics[width=0.2\textwidth]{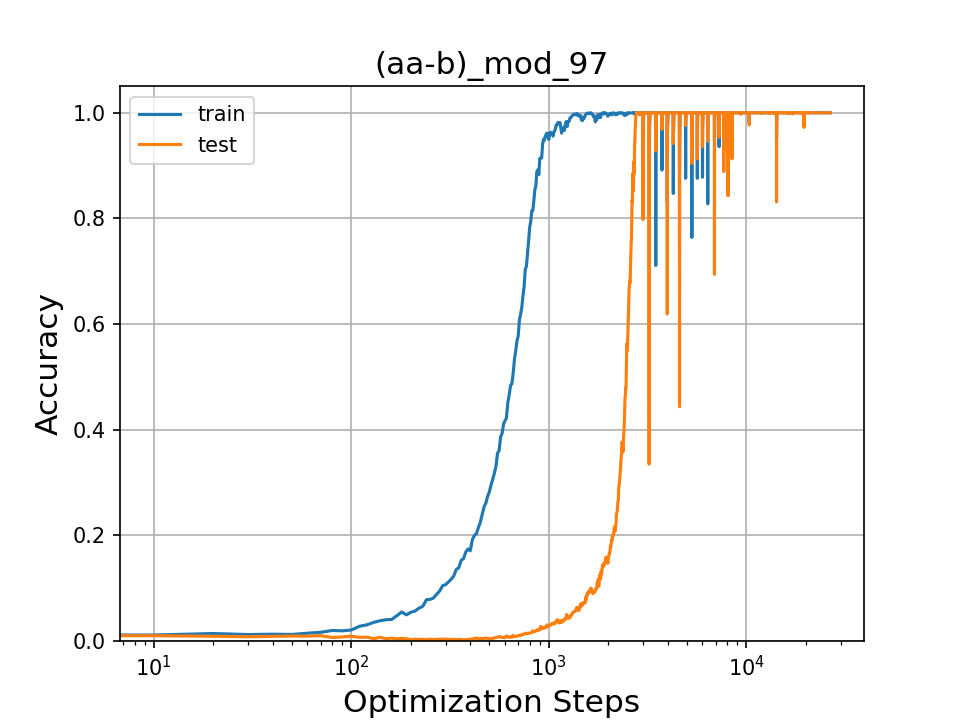}}
  \hfill
  \subfloat[\texttt{axc+bxd-e (mod 7)}.]{\includegraphics[width=0.2\textwidth]{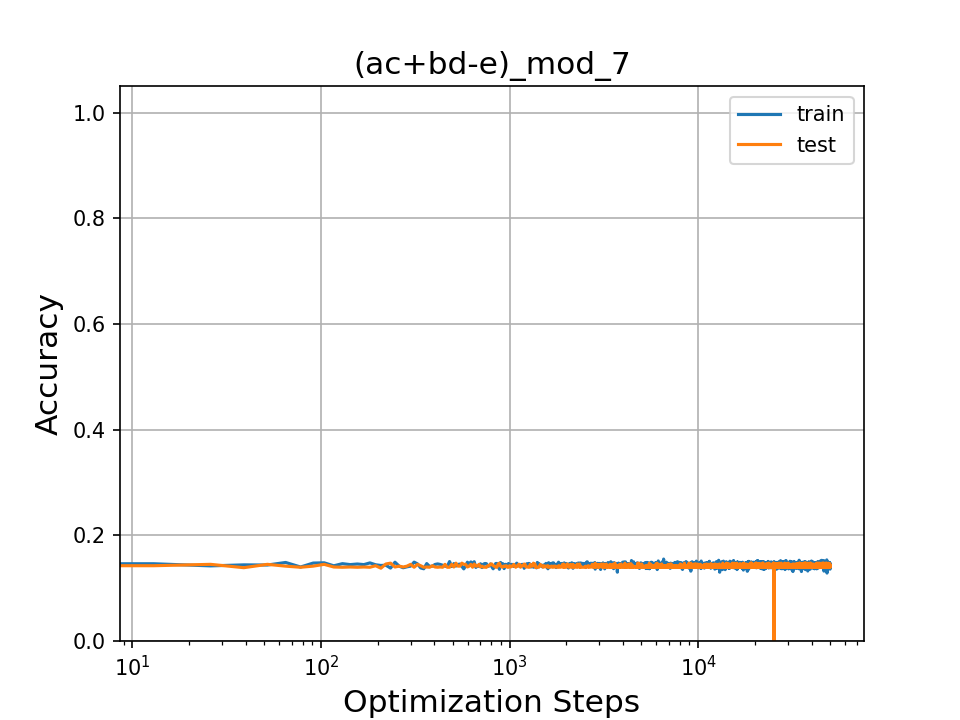}}
  \caption{\textsc{Grokfast-EMA} results on all tasks.}
  \label{appd:fig:grokfast-ema}
\end{figure}

\citet{lee2024grokfastacceleratedgrokkingamplifying} also propose another version named \textsc{Grokfast-EMA}. We follow the hyperparameter settings recommended in their original paper, and test the performances on all five tasks. 
The results are shown in \autoref{appd:fig:grokfast-ema}. We can observe the instability from \textsc{Grokfast-EMA}, which is also sensitive to hyperparameters. As the standard training baseline, it also fails in the hardest task.

\newpage
\section{Compare Weight-decay Regularization and Gradient Rescaling on the Challenging Task}
\subsection{Task5: (\texttt{axc+bxd-e}) mod 97}
\begin{figure}[!h]
      \centering
	   \subfloat[No Regularization.]
		{\includegraphics[width=0.48\textwidth]{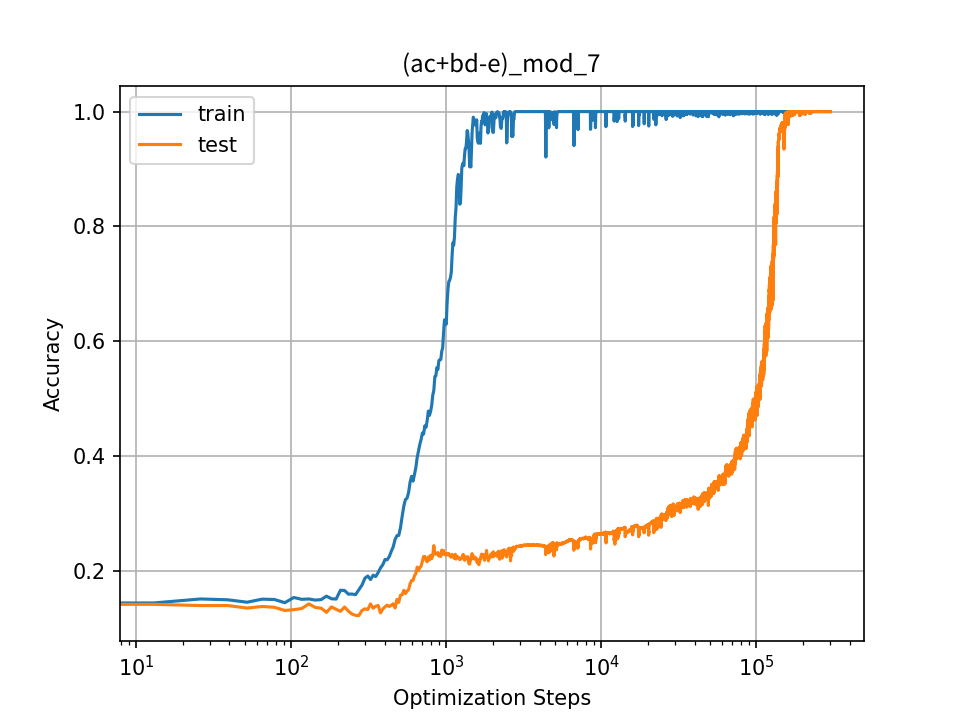}}
        \hfill
	   \subfloat[With Standard Gradient Normalization.]
		{\includegraphics[width=0.48\textwidth]{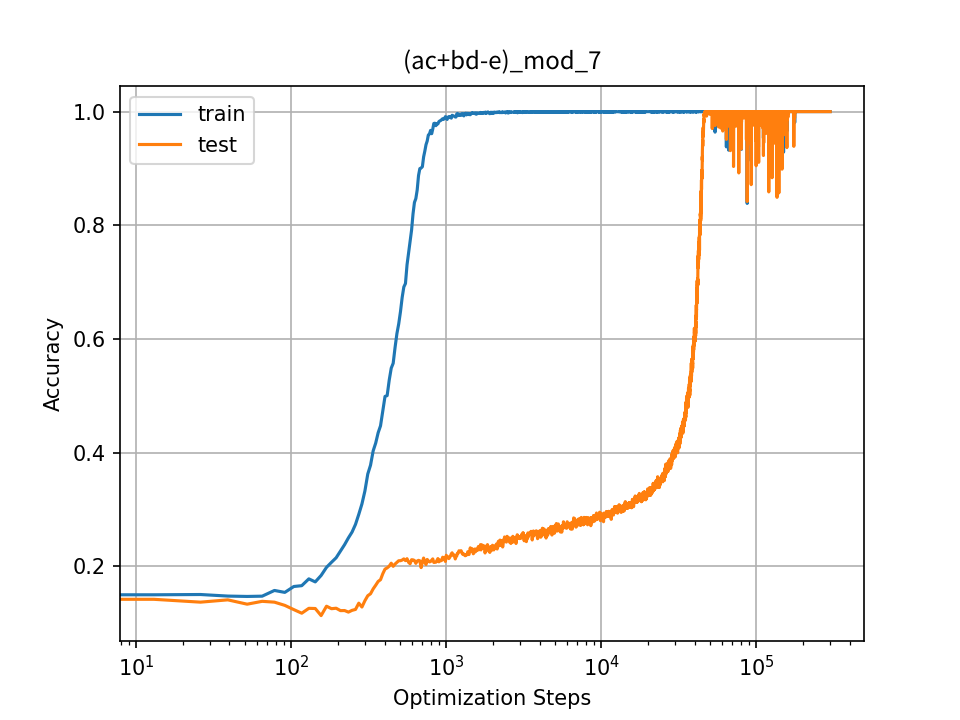}}
        \quad
      \subfloat[With Weight-decay.]
		{\includegraphics[width=0.48\textwidth]{fig_prepare/acc_Nonormed_ac_add_bd_sub_e_mod_7_p=7_TD_4Layers_4Heads_128Dim_lr0.001_wd0.001.png}}
        \hfill
        \subfloat[With Weight-decay and Standard Gradient Normalization.]
		{\includegraphics[width=0.48\textwidth]{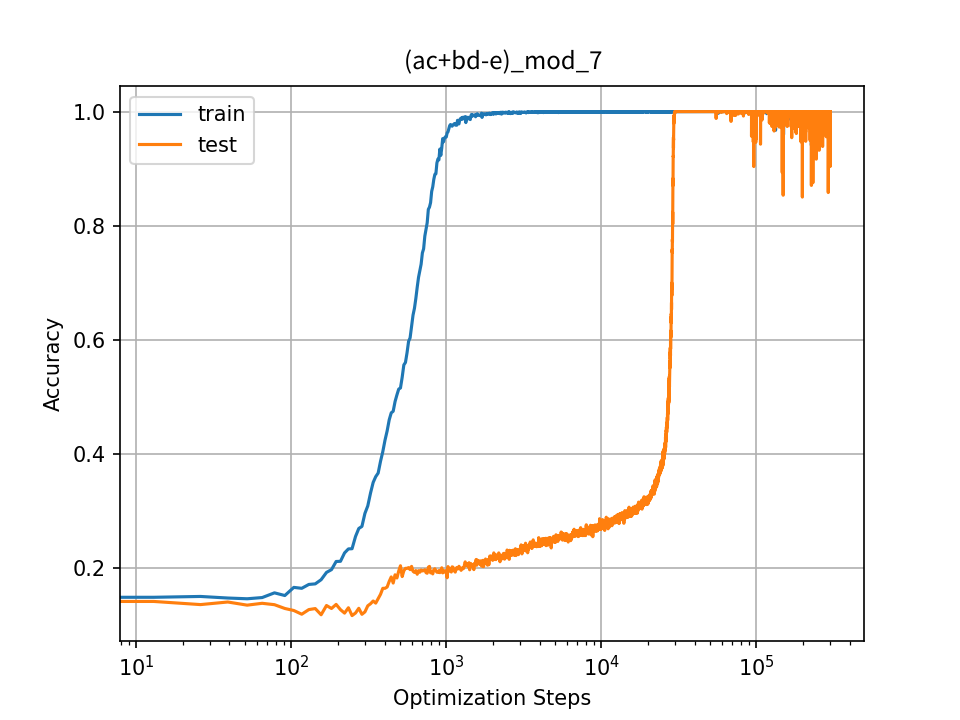}}
	\caption{\textbf{Standard training on $ac+bd-e\ (\text{mod }7)$ with various combination of regularization techniques.} }
	\label{fig:std-wd-rescale-ablation-ac+bd-e}
\end{figure}

We compare the effects of conventional weight-decay regularization and gradient rescaling on $ac+bd-e\ (\text{mod }7)$ in \autoref{fig:std-wd-rescale-ablation-ac+bd-e}. Only applying the standard gradient normalization can effectively stabilize the training, but leads to a larger gap between overfitting and generalization under the grokking phenomenon. We recommend to apply a small value of weight decay with standard gradient normalization to achieve the best performance on challenging arithmetic tasks.

\newpage
\section{Transferrability of the Neural Amplifier between Different Arithmetic Tasks}
\label{appd:fig:transfer-task1-2}
While \textsc{NeuralGrok} accelerates grokking within individual arithmetic tasks, we find that the learned gradient transformations exhibit limited transferability even across operations leveraging similar correlations between variables and operators (e.g., $+$ vs. $-$). 

\begin{figure}[!ht]
  \centering
  \subfloat[$a+b\  (\text{mod }97)$ to $a-b\ (\text{mod }97)$.]{\includegraphics[width=0.24\textwidth]{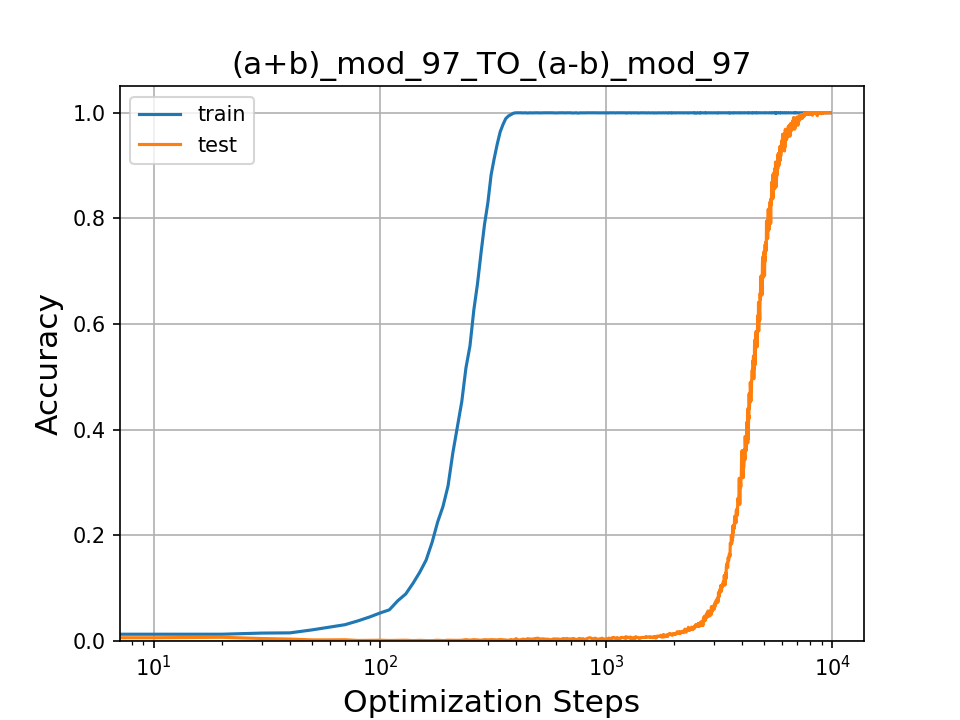}}
  \hfill
  \subfloat[$a\cdot b\  (\text{mod }97)$ to $a- b\ (\text{mod }97)$.]{\includegraphics[width=0.24\textwidth]{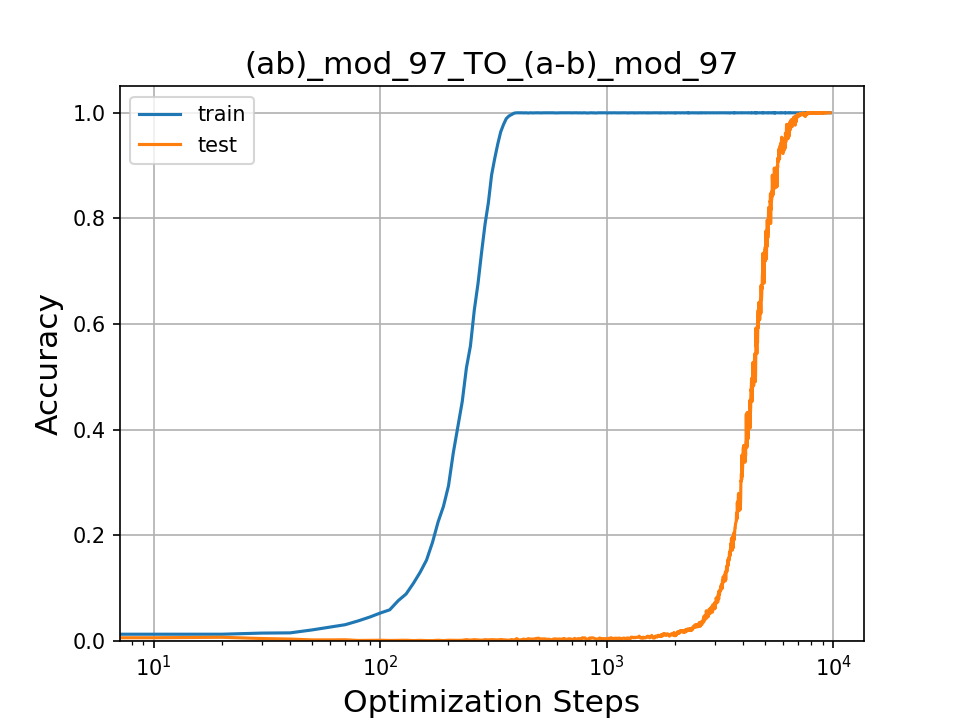}}
  \hfill
  \subfloat[$a\times a-b\  (\text{mod }97)$ to $a b\ (\text{mod }97)$.]{\includegraphics[width=0.24\textwidth]{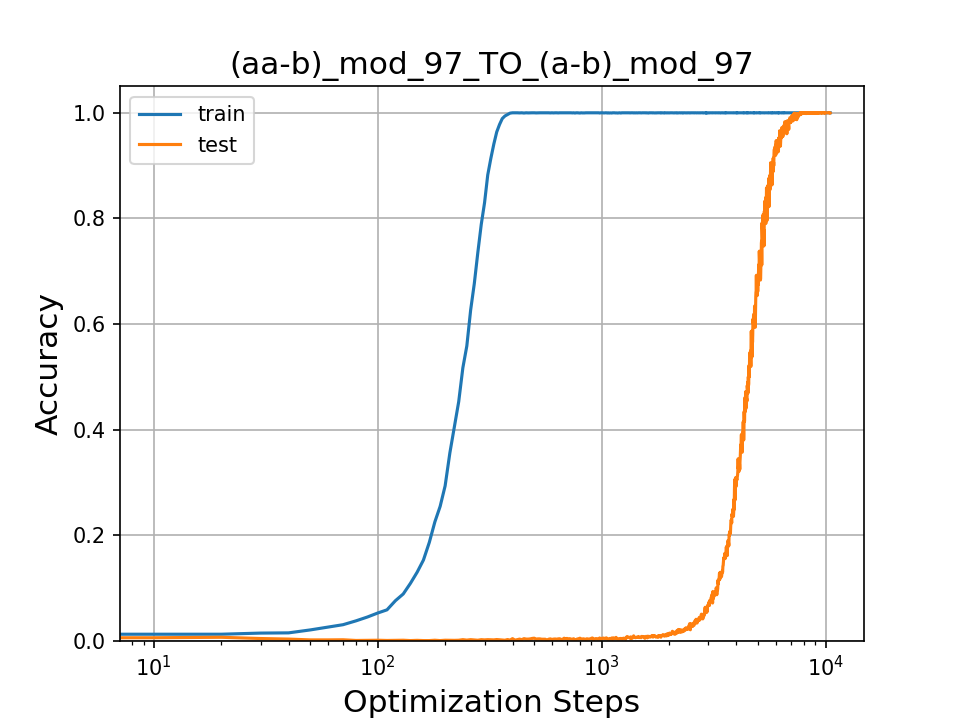}}
  \hfill
  \subfloat[$ac+bd-e\ (\text{mod }7)$ to $a-b \ (\text{mod }97)$.]{\includegraphics[width=0.24\textwidth]{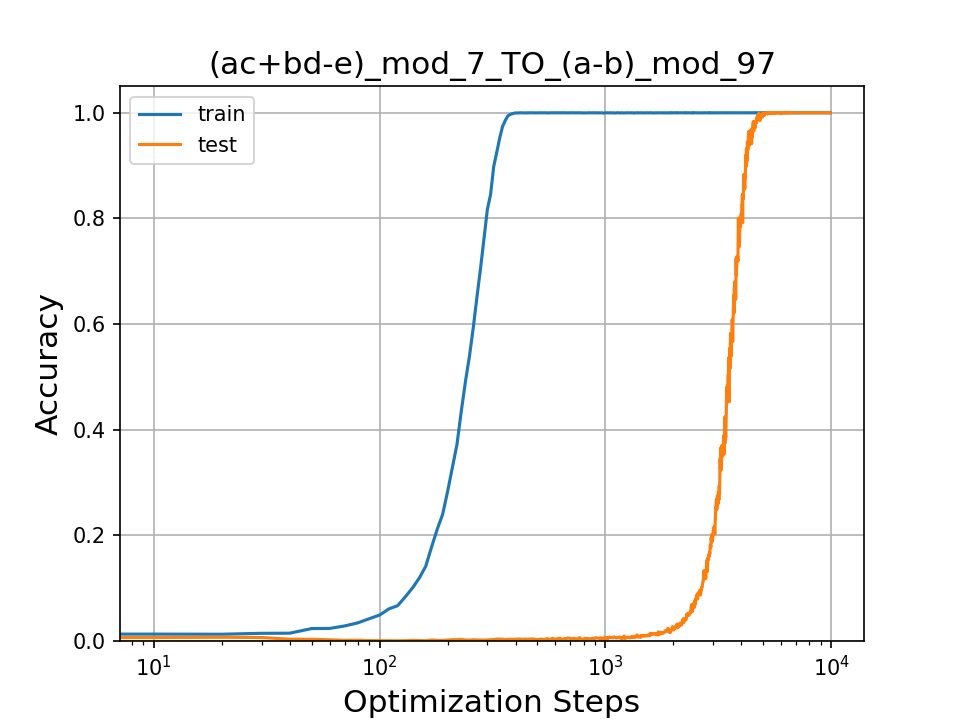}}
  \caption{Transfer learning experiments from other tasks to $a-b\ (\text{mod }97)$. There is still an obvious gap between memorization and generalization, even when previously pretrained on the hardest task. }
  \label{fig:(a-b)-transfer}
\end{figure}

This suggests that the \textit{neural-amplifier} adapts to highly task-specific gradient patterns, such as suppressing noise in cyclic modular operations or amplifying critical features to disentangle composite equations. 
For instance, transformations optimized for modular addition ($+$) may fail to generalize to subtraction ($-$) or mixed-operation tasks (\autoref{fig:(a-b)-transfer}), where gradient dynamics not only cares about correlations between variables and operators, but also involve nuance reasoning mechanisms. 
This specialization might stem from the interplay between the bilevel optimization framework and the narrow validation objectives ($\mathcal{D}_{outer}$), which force the amplifier to local task geometries rather than global arithmetic principles. Future work could investigate cross-task meta-learning or shared amplification modules to disentangle universal arithmetic patterns from task-specific adaptations.

\end{document}

%% file: math_commands.tex
\usepackage{amsmath,amsfonts,bm}

\def\eqref#1{equation~\ref{#1}}

\def\1{\bm{1}}

\def\vtheta{{\bm{\theta}}}

\def\vg{{\bm{g}}}

\def\vp{{\bm{p}}}

\DeclareMathAlphabet{\mathsfit}{\encodingdefault}{\sfdefault}{m}{sl}
\SetMathAlphabet{\mathsfit}{bold}{\encodingdefault}{\sfdefault}{bx}{n}

\DeclareMathOperator*{\argmin}{arg\,min}